\documentclass[lettersize,journal]{IEEEtran}
\usepackage{amsmath,amsfonts}
\usepackage{algorithmic}
\usepackage{algorithm}
\usepackage{array}
\usepackage[caption=false,font=normalsize,labelfont=sf,textfont=sf]{subfig}
\usepackage{hyperref}
\usepackage{textcomp}
\usepackage{stfloats}
\usepackage{url}
\usepackage{verbatim}
\usepackage{graphicx}
\usepackage{cite}
\usepackage{booktabs}
\usepackage{xcolor}
\usepackage{graphicx}
\usepackage{multirow} 
\begin{document}

\title{Back2Color: Domain-Adaptive Synthetic-to-Real Monocular Depth Estimation for Dynamic Traffic Scenes}

\author{Yufan~Zhu, Chongzhi~Ran, Mingtao~Feng, 
Le~Dong,~\IEEEmembership{Member,~IEEE},
Weisheng~Dong,~\IEEEmembership{Member,~IEEE}, Antonio~M.~López,~\IEEEmembership{Member,~IEEE}

\thanks{Yufan Zhu, Chongzhi Ran, Mingtao Feng, Le Dong, Weisheng Dong are with the School of Artificial Intelligence Xidian University, Xi’an 710071, China(e-mail:zhuyufan@xidian.edu.cn; czran@stu.xidian.edu.cn;  mintfeng@hnu.edu.cn; dongle@xidian.edu.cn; wsdong@mail.xidian.edu.cn; gmshi@xidian.edu.cn).}
\thanks{Antonio M. López is with the Department of Computer Science and also with the Computer Vision Center (CVC), Universitat Autònoma de Barcelona (UAB), 08193 Bellaterra (Barcelona), Spain(e-mail:antonio@cvc.uab.cat).}
}



\maketitle

\begin{abstract}
Accurate monocular depth estimation is a fundamental component of vision-based perception systems in intelligent transportation applications. Despite recent progress, unsupervised monocular approaches still suffer from significant performance degradation in real-world traffic scenes due to synthetic-to-real domain gaps and the presence of dynamic, non-rigid objects such as vehicles and pedestrians. In this paper, we propose Back2Color, a robust unsupervised monocular depth estimation framework that addresses these challenges through domain adaptation and uncertainty-aware fusion. Specifically, Back2Color proposes a bidirectional depth-to-color transformation strategy that learns appearance mappings from real-world driving data and applies them to synthetic depth maps, thereby constructing training samples with realistic color appearance and paired synthetic depth. In this way, the proposed approach effectively reduces the domain gap between simulated and real traffic scenes, enabling the depth prediction network to learn more stable and generalizable priors. To further improve robustness under dynamic environments, we propose an auto-learning uncertainty temporal-spatial fusion (Auto-UTSF) module, which adaptively fuses complementary temporal and spatial cues by estimating pixel-wise uncertainty, enabling reliable depth prediction in the presence of moving objects and occlusions. Extensive experiments on challenging urban driving benchmarks, including KITTI and Cityscapes, demonstrate that the proposed method consistently outperforms existing unsupervised monocular depth estimation approaches, particularly in dynamic traffic scenarios, while maintaining high computational efficiency.

\end{abstract}


\begin{figure}[t]
    \centering
        \includegraphics[width=3.4in]{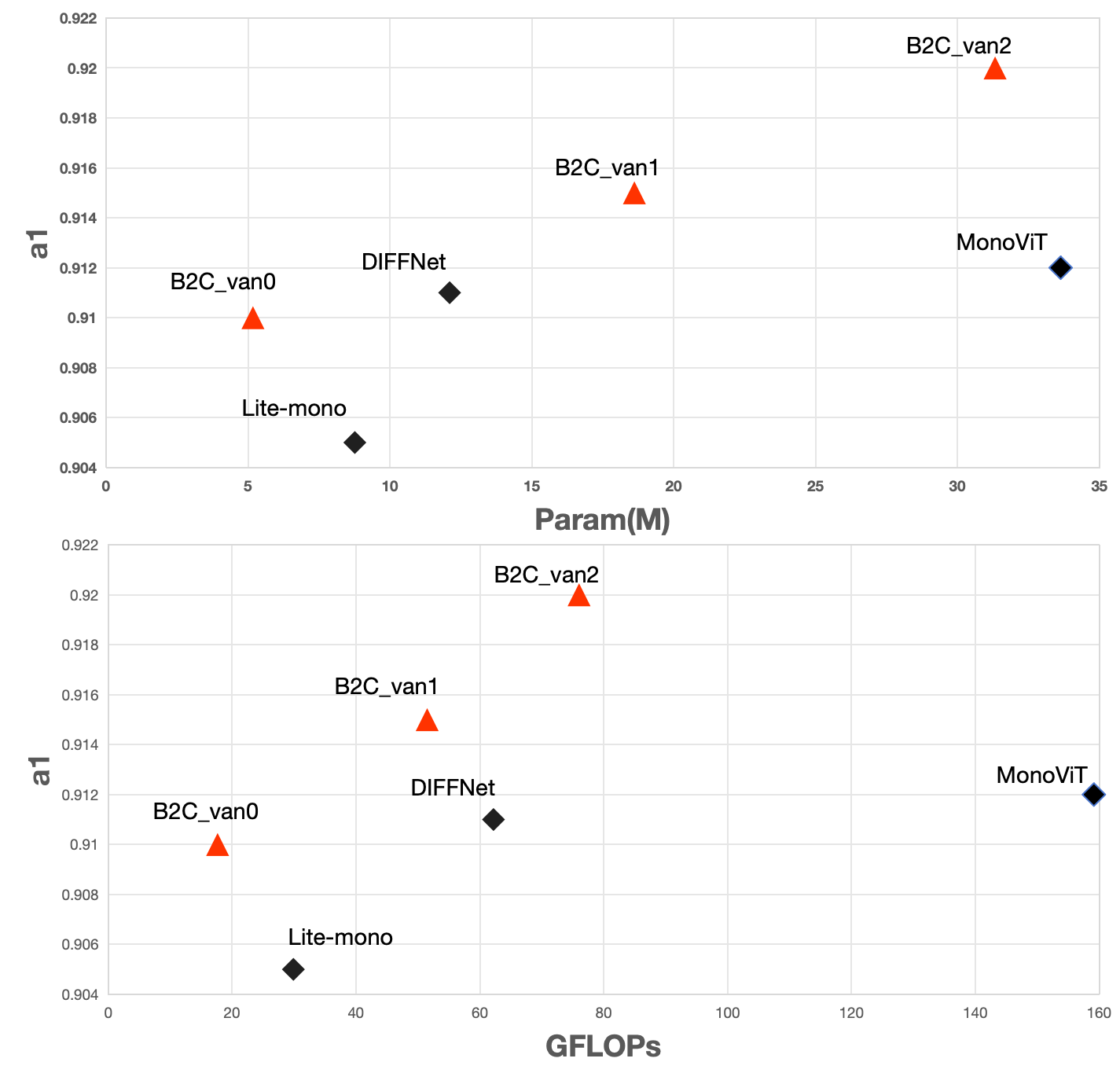}
        \vspace{-3mm}
        \caption{Performance vs. methods efficiency on KITTI at a resolution of 320 × 1024. Our method(B2C) surpasses both the lightweight Lite-mono\cite{zhang2023lite} and the highly effective MonoViT\cite{zhao2022monovit}, excelling in both efficiency and accuracy.} 
        \label{fig:show}
        \vspace{-4mm}
\end{figure}

\section{Introduction}
\IEEEPARstart{I}{n} the real world, even after prolonged evolution, humans are still limited to qualitatively analyzing the depth of specific points in their field of view to assist in their actions\cite{hochberg1952familiar}. Despite significant technological advances, the acquisition of precise and dense depth information remains a formidable challenge\cite{1384826,li2022progress,kim2018deep,8016631,lin2013absolute,li2023temporally}. Even with the refinement of advanced LiDAR systems\cite{li2022progress}, their data are inherently sparse, comprising only discrete depth points. Unsupervised depth estimation(UDE), powered by triangulation\cite{luong1996fundamental} using purely visual approaches does not require additional annotation for depth information other than camera parameters. UDE eliminates the necessity of annotating depth labels for specific scenes. By utilizing binocular and motion disparity clues, it mimics the structure of human vision, making it efficient and cost-effective\cite{blake2011binocular}. This is why many researchers and companies are actively pursuing this approach\cite{ingle2016tesla}. Monodepth2\cite{godard2019digging}, the most renowned framework for UDE, has established itself as a classic and successful method. By reprojecting related frames(temporal and spatial) and regressing depth values. This adaptability enables Monodepth2 to obtain depth information for various scenes through adaptive fine-tuning, making it highly versatile. However, despite its strengths, Monodepth2 has several limitations. Firstly, the auto-masking for handling non-rigid moving objects, follows the rule: the pixel error after reprojection should be less than before reprojection. This pixel-level determination cannot actively detect motion\cite{10321683} and cannot ensure effectiveness in smooth regions. Additionally, when applied to more complex datasets such as Cityscapes\cite{cordts2016cityscapes}, it almost fails. Secondly, photometric reprojection error also fails in areas with no texture or with texture-less background objects, leading to depth prediction failure. Lastly, methods based on photometric reprojection matching lack prior knowledge of the objective structure. These issues make it challenging for UDE methods to be applied in real world.

\par With the continuous advancement of computer graphics and virtual engine technologies, significant progress has been made in the creation and application of virtual urban scene datasets\cite{cabon2020vKitti2,ros2016synthia,cordts2016cityscapes,huang2018apolloscape,dosovitskiy2017carla}. The evolution of 3D virtual engines, such as Unity and Unreal Engine\cite{haas2014history,sanders2016introduction}, has played a crucial role in enhancing the realism and visualization effects of virtual urban scenes. These engines not only facilitate finer scene modeling but also provide interactivity and dynamism, making virtual urban scene data more closely resemble the real world. Synthetic datasets can provide precise qualitative data for virtual urban scenes, including depth, semantic labels, and more. Therefore, it would be desirable if various synthetic scenes and dense depth could be used to assist in the UDE of the real-world\cite{gurram2021monocular}. Compared to synthetic datasets, real-world depth estimation datasets\cite{geiger2012we,cordts2016cityscapes} have the following notable drawbacks: firstly, synthetic depth data is dense, whereas depth data collected by LiDAR is sparse and discrete\cite{geiger2012we}. Secondly, depth computed through stereo matching suffers from occlusions and lacks accuracy due to the short baseline\cite{cordts2016cityscapes}. Additionally, sparse depth sampling does not provide sufficient information for learning when obstacles are small or distant (resulting in small apparent areas), making it challenging for the network to accurately reconstruct their shapes.

\begin{figure*}[!t]
    \centering
        \includegraphics[width=7.1in]{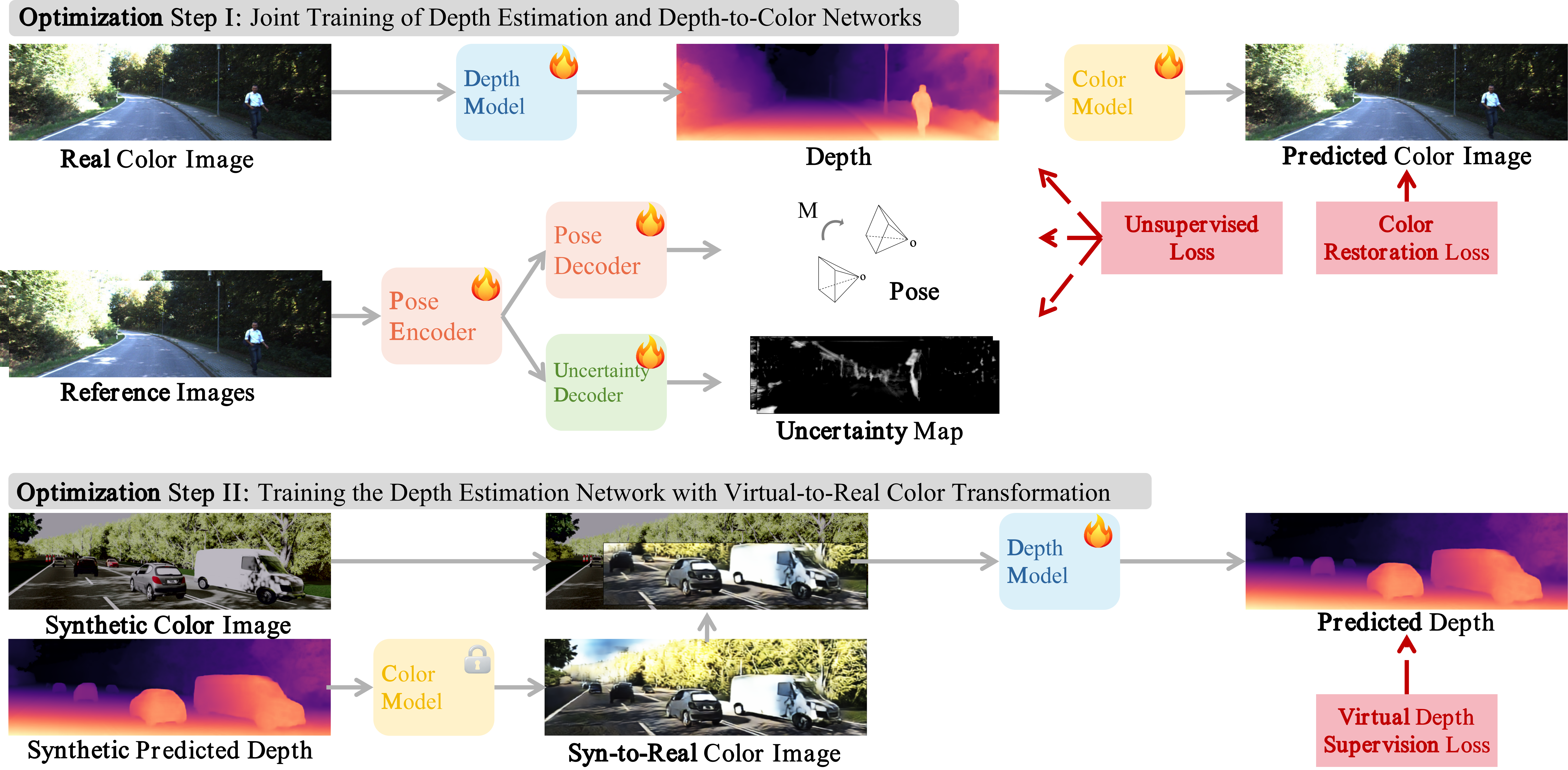}
        \caption{\textbf{Illustration of the Back2Color training strategy.} In each iteration, the first stage jointly optimizes the depth estimation network and the depth-to-color network using real-world data with unsupervised photometric and color restoration losses. In the second stage, the depth-to-color network is fixed and applied to transform synthetic depth maps into realistic color images, enabling the depth estimation network to be trained with virtual depth supervision under a reduced domain discrepancy.} 
        \label{fig:framework} 
\end{figure*}

\par In this paper, we propose the Back2Color framework to address the appearance discrepancy between synthetic and real-world data in unsupervised monocular depth estimation. Specifically, Back2Color introduces a bidirectional depth-to-color transformation strategy that learns appearance mappings from real-world driving scenes and applies them to synthetic depth maps. In this way, Back2Color constructs training pairs that combine realistic color appearance with corresponding synthetic depth, effectively narrowing the domain gap between simulated and real traffic scenes. 

To further enable effective joint training across heterogeneous data sources, we introduce an enhanced CutMix strategy, termed Syn-Real CutMix~\cite{yun2019cutmix}, which mixes synthetic and syn-to-real images at both batch and spatial levels. This strategy enriches the training data with diverse color–depth associations derived from synthetic scenes while preserving real-world appearance characteristics. By jointly training on these complementary data sources, the proposed framework allows the depth estimation network to learn more stable and generalizable depth prediction priors, leading to improved robustness in real traffic environments.

In addition, we observe that existing auto-masking mechanisms often fail to reliably identify non-rigid moving objects in traffic scenes, especially when vehicles move at similar speeds or in opposite directions, which leads to erroneous depth predictions. To address this issue, we introduce an automatic uncertainty learning fusion strategy that combines temporal and spatial cues, termed Auto-learning Uncertainty Temporal-Spatial Fusion (Auto-UTSF)~\cite{zhu2024tsudepth}. By explicitly modeling uncertainty, Auto-UTSF adaptively integrates complementary information from stereo and temporal supervision, enabling more reliable depth estimation under dynamic and occluded conditions.

To summarize, the main contributions of this work are as follows:

\begin{itemize}
\item We propose Back2Color, a synthetic-to-real adaptation framework that learns scene-specific color information from real-world data and enables reverse color prediction from depth. Together with the proposed Syn-Real CutMix strategy, Back2Color effectively harmonizes joint training on synthetic and real datasets, improving the robustness of unsupervised monocular depth estimation in real traffic scenes.
\item We introduce Auto-learning Uncertainty Temporal-Spatial Fusion (Auto-UTSF), which robustly handles non-rigid motion and occlusions by adaptively fusing complementary spatial and temporal supervision through uncertainty estimation.
\item Extensive experiments on challenging urban driving benchmarks demonstrate that the proposed method consistently improves depth estimation accuracy while maintaining computational efficiency, and produces depth maps with clearer object boundaries.
\end{itemize}

\begin{figure*}[t]
    \centering
        \includegraphics[width=7.1in]{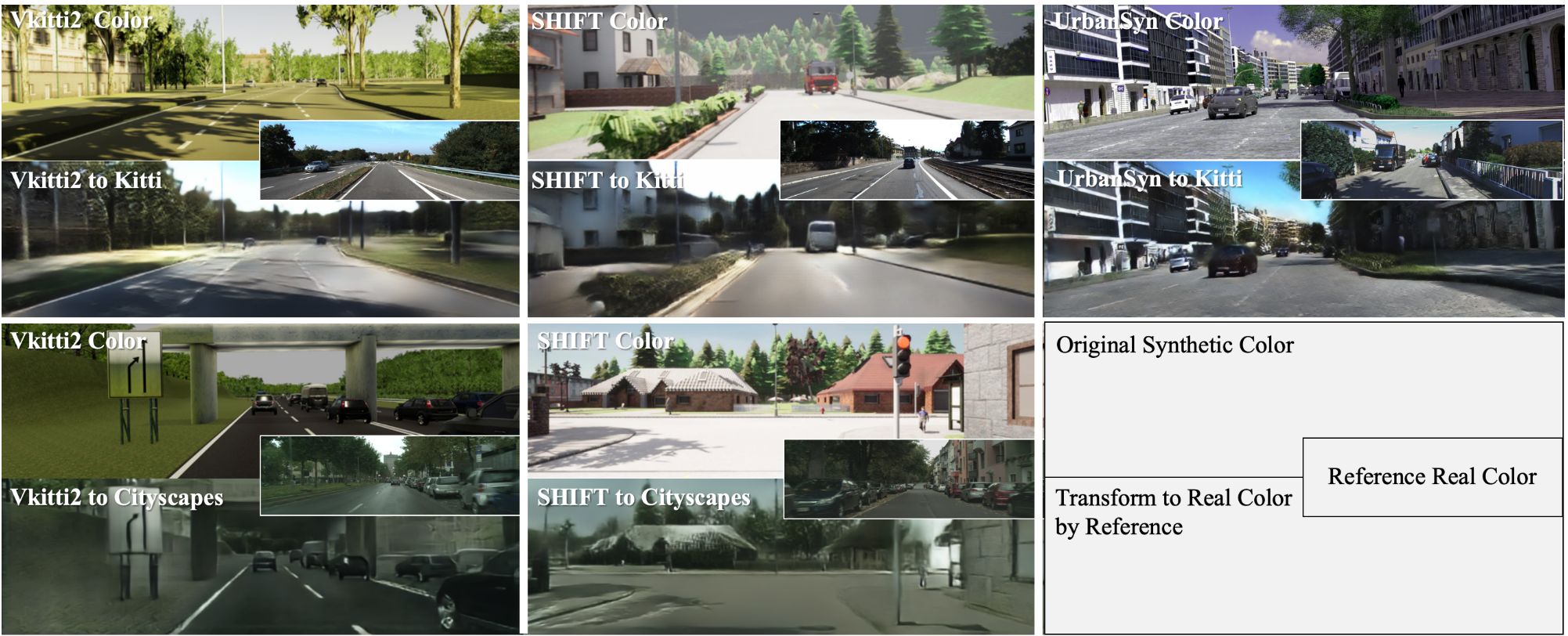}
        \vspace{-5mm}
        \caption{\textbf{Color transformation for specific datasets}: we showcase the capability of our Back2Color framework to effectively transform various synthetic datasets, including vKITTI2\cite{cabon2020vKitti2}, SHIFT\cite{shift2022}, and UrbanSyn\cite{gómez2023one}, into Syn2Real images. These images closely emulate the real-world styles of KITTI\cite{eigen2014depth,geiger2012we} and Cityscapes\cite{cordts2016cityscapes}, demonstrating high fidelity and credibility.} 
        \label{fig:examples_back2Color}
        \vspace{-3mm}
    \end{figure*}

\section{Related Works}

\label{sec:related works}
\subsection{Unsupervised Depth Estimation}
    Unsupervised depth estimation aims to predict distances from stereo views or video sequences. Initial methods like those by Godard et al.\cite{godard2017unsupervised} introduced disparity consistency loss, while Zhou et al.\cite{zhou2017unsupervised} pioneered depth and pose estimation simultaneously. Recent advancements\cite{guizilini20203d, yan2021channel, lyu2021hr, zhou2021self, he2022ra,10570231} refine network structures for higher accuracy. Techniques like PackNet\cite{guizilini20203d} and CADepth\cite{yan2021channel} enhance geometric and appearance information propagation. HR-Depth\cite{lyu2021hr} and DIFFNet\cite{zhou2021self} utilize HRNet for better resolution utilization. RA-Depth\cite{he2022ra} employs HRNet for both encoding and decoding. In natural language processing, the Transformer architecture\cite{vaswani2017attention} has been successful, later adapted to computer vision tasks\cite{dosovitskiy2020image, liu2021swin}. It serves as the backbone for depth estimation\cite{ranftl2021vision, varma2022transformers, zhao2022monovit, zhang2023lite, lee2022mpvit}, but faces challenges like computational demands and real-time constraints. Existing methods prioritize model architecture, often overlooking crucial obstacles and targets. Additionally, depth estimation models also face the issue of overfitting on specific datasets and a lack of continuous learning capability\cite{zhang2024reusable}.

\subsection{Depth Estimation with Synthetic data}

    \par Researchers have already attempted to use synthesized dense depths to guide the training of real-world depth estimation\cite{swami2022you,gurram2021monocular}. Gurram et al.\cite{gurram2021monocular} categorized this problem as a dimension adaptation problem, where the objective is to adapt models trained on a source(synthesized) domain to perform well in a different target domain(real-world). They employed the Gradient Reverse Layer\cite{ganin2015unsupervised} to identify features from different domains during back propagation. Swami et al.\cite{swami2022you}, on the other hand, simultaneously trained on both synthesized and real-world datasets, leveraging scale information in Synthetic dataset to recover scale in unsupervised depth estimation.
    To bridge the gap between real-world and synthetic datasets in the image domain, Zhao et al.\cite{zhao2020domain} leveraged synthetically rendered data and employed image translation techniques to narrow the synthetic-real domain gap, thereby enhancing monocular depth prediction. 

\subsection{Style transfer in urban scenes}
    Research on style transfer for urban scene images has long been ongoing due to its importance in autonomous driving technology. Isola et al.\cite{isola2017image} proposed a conditional GAN framework for generating urban scene images from semantic segmentation images of Cityscapes in 2017. However, generative models may produce random and plausible objects, disrupting the scene structure, as shown in their experiments. Atapour et al.\cite{atapour2018real} utilized style transfer and adversarial training to convert real-world color images into synthetic environment-style images. Huang et al.\cite{huang2018auggan} applied GAN-based domain transfer for data augmentation in urban scene weather-time conversion. To minimize the gap between source and target domains, Gomez et al.\cite{gomez2023co} aligned colors across different domains in the Lab color space. Zheng et al.\cite{zheng2018t2net} used GANs to convert synthetic colors to real-world colors. 
    With advancements in stable diffusion\cite{zhang2023adding,upadhyay2023enhancing}, ControlNet \cite{zhang2023adding} is now capable of generating images based on various conditions, including text, pose, Canny edges, and depth. This capability theoretically enables the generation of RGB images from urban depth maps, as demonstrated by RISHI et al. \cite{upadhyay2023enhancing}. However, experiments \cite{upadhyay2023enhancing} indicate that diffusion models do not strictly adhere to depth information and may arbitrarily alter target classes. Additionally, diffusion models require 2-5 seconds to generate a single image espically trainning with CNN model, and each condition demands extensive data for training. This makes it challenging to quickly fine-tune a depth estimation model for specific real-world scenarios. In contrast, our proposed ColorNet addresses these limitations, allowing for rapid deployment and real-time operation.
    
\section{Methodology}
In this section, we present the Back2Color framework, which captures the depth-to-color mapping within the target domain and transfers synthetic colors to pseudo-realistic ones (\ref{section:depth2colors}). Back2Color is further enhanced with Real-Syn CutMix (\ref{section:Syn-Real CutMix}) to jointly train for unsupervised depth estimation in the real world and supervised depth estimation in synthetic data. Then, we describe the Auto-learning Uncertainty Temporal-Spatial Fusion (Auto-UTSF) module (\ref{section:Auto-UTSF}), designed to address challenges associated with non-rigid motion. Throughout this paper, Depth and Disparity are considered equivalent.

\begin{figure*}[!t]
    \centering
        \includegraphics[width=7.1in]{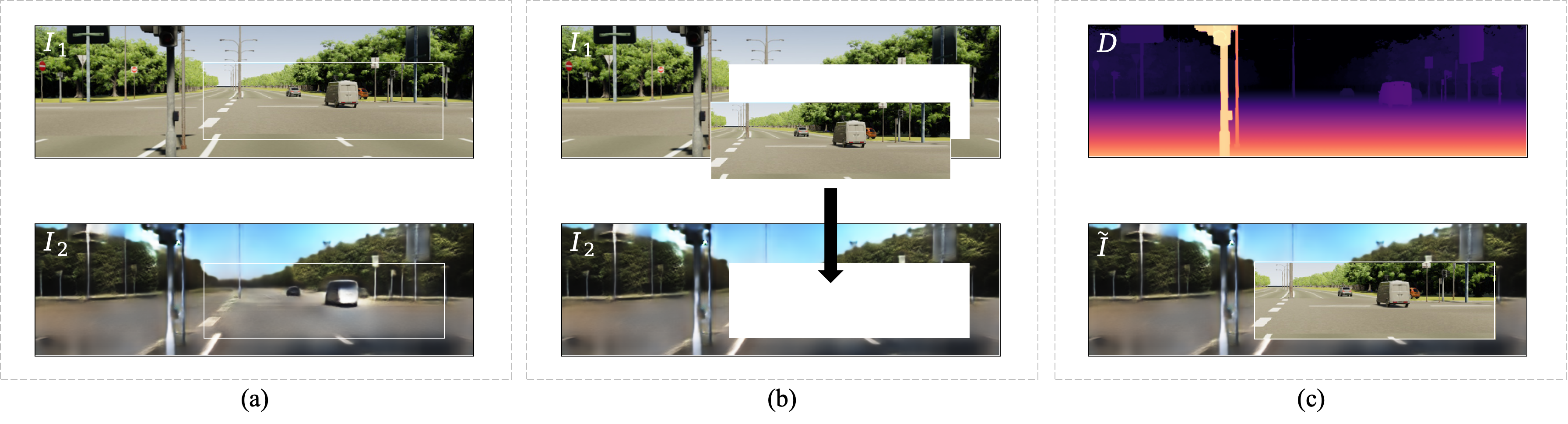}
        \vspace{-6mm}
        \caption{\textbf{Syn-Real CutMix}: specific steps involved in our Syn-Real CutMix method for spatially combining colors: (a) Randomly select a rectangular region, denoted as M. (b) Replace the corresponding block in image $I_2$ with the block from image $I_1$. (c) Resulting in the formation of new paired data $(\tilde{I}, D)$.} 
        \label{fig:Syn_Real_CutMix}
        \vspace{-5mm}
\end{figure*}

\subsection{Back to the Color}
\subsubsection{Transform depth to specific colors}
\label{section:depth2colors}
\par the significant gap between synthetic and real-world color images arises from the inherent complexity of the real world. Colors in real-world scenes are influenced by various factors, including lighting intensity, surface material of obstacles, and smoothness, and can also be significantly degraded\cite{10130799}. Each of these factors is intricate and highly variable, making them challenging to simulate realistically using typical 3D engines. As a result, discerning the origin of an image intuitively is easy, both for humans and deep learning alike. Consequently, directly combining real-world and synthetic datasets for training essentially involves dealing with the integration issues of two distinct domains. Our Back2Color approach learns the mapping relationship in the target domain(Real). This mapping is then applied to the source domain (Synthetic) to reduce the gaps between colors of the source and target domains.
    
\par Objectively speaking, predicting color from 3D structural information constitutes a one-to-many mapping, a non-standard conditional generative tasks, where the same 3D structure can correspond to a multitude of different color images. However, under specific conditions, this seemingly impossible task becomes feasible: \textit{Assuming that the color images in the target domain exhibit relatively high correlation.} The colors in real word datasets like Cityscapes and KITTI are notably different, as shown in Fig.\ref{fig:examples_back2Color}. Therefore, in order to bridge the substantial gap between synthetic $\mathcal{S} \in$ \{vKITTI2(vK)\cite{cabon2020vKitti2}, SHIFT(SH)\cite{shift2022}, UrbanSyn(Ur)\cite{gómez2023one}\} and real-world $\mathcal{X} \in$ \{KITTI(KT)\cite{eigen2014depth,geiger2012we}, Cityscapes(Cs)\cite{cordts2016cityscapes}\} datasets, we can learn the depth-to-color mapping in the target domain (real world color) and apply it to the source domain (synthetic color), thereby converting the source domain color to match those in the target domain. To bridge the appearance gap between synthetic and real-world data, we train a depth-to-color network $\mathcal{F}_{d2c}$ that predicts color images from estimated depth maps in the real-world domain. Given a predicted depth map $\hat{\mathbf{D}}_{\mathcal{X}}$ produced by the depth estimation network, the depth-to-color network reconstructs the corresponding color image as
\begin{equation}
\hat{\mathbf{I}}_{\mathcal{X}} = \mathcal{F}_{d2c}(\hat{\mathbf{D}}_{\mathcal{X}};\boldsymbol{\theta}_{d2c}),
\label{eq:back2color1}
\end{equation}
where $\boldsymbol{\theta}_{d2c}$ denotes the learnable parameters of the depth-to-color network.

The reconstructed color image $\hat{\mathbf{I}}_{\mathcal{X}}$ is supervised by the corresponding real color image $\mathbf{I}_{\mathcal{X}}$ using a photometric reconstruction loss, defined as
\begin{equation}
\mathcal{L}_{d2c} = pe(\hat{\mathbf{I}}_{\mathcal{X}}, \mathbf{I}_{\mathcal{X}}),
\label{eq:back2color2}
\end{equation}
where $pe(\cdot,\cdot)$ denotes the photometric error function.

Next, we investigate whether the depth-to-color mapping learned from the real-world domain can be transferred to synthetic data. Specifically, after training the depth-to-color network $\mathcal{F}_{d2c}$ on real-world samples, we apply the learned model to predicted depth maps from the synthetic domain in order to generate color images that resemble the appearance characteristics of the target domain (e.g., KITTI), as illustrated in Fig.~\ref{fig:examples_back2Color}. Formally, given a predicted synthetic depth map $\hat{\mathbf{D}}_{\mathcal{S}}$, the corresponding color image is generated as
\begin{equation}
\hat{\mathbf{I}}_{\mathcal{S}\rightarrow \mathcal{X}} = \mathcal{F}_{d2c}(\hat{\mathbf{D}}_{\mathcal{S}};\boldsymbol{\theta}_{d2c}),
\label{eq:back2color3}
\end{equation}
where $\hat{\mathbf{I}}_{\mathcal{S}\rightarrow \mathcal{X}}$ denotes the synthetic color image transformed to match the appearance of the real-world domain.

\subsubsection{Syn-Real CutMix}
\label{section:Syn-Real CutMix}
\par The objective of CutMix in our framework is to construct enhanced training pairs $(\tilde{\mathbf{I}}, \mathbf{D})$ for improving robustness across domains. Unlike the original CutMix~\cite{yun2019cutmix}, which mixes two independent image--label pairs, the image pair used in our setting, $(\mathbf{I}_{\mathcal{S}}, \hat{\mathbf{I}}_{\mathcal{S}\rightarrow\mathcal{X}})$, represents two different color appearances corresponding to the same synthetic depth label $\mathbf{D}_{\mathcal{S}}$. Therefore, we perform mixing only in the color space while keeping the depth supervision unchanged.

\par To introduce spatial diversity, as illustrated in Fig.~\ref{fig:Syn_Real_CutMix}, we denote the image pair $(\mathbf{I}_{\mathcal{S}}, \hat{\mathbf{I}}_{\mathcal{S}\rightarrow\mathcal{X}})$ as $(\mathbf{I}_1, \mathbf{I}_2)$, which are randomly swapped with equal probability. A rectangular region with height and width larger than half of the original image is then randomly cropped from $\mathbf{I}_1$ and replaced with the corresponding region from $\mathbf{I}_2$, yielding the mixed image
\begin{equation}
\tilde{\mathbf{I}} = \mathbf{M} \odot \mathbf{I}_1 + (1 - \mathbf{M}) \odot \mathbf{I}_2,
\label{eq:CutMix}
\end{equation}
where $\mathbf{M} \in \{0,1\}^{H \times W}$ denotes a binary mask indicating the preserved and replaced regions, and $\odot$ represents element-wise multiplication.

\par The Syn-Real enhanced images $\tilde{\mathbf{I}}$ are then combined with real-world target images $\mathbf{I}_{\mathcal{X}}$ within the same training batch. This joint batching strategy allows the monocular depth estimation network to be trained simultaneously on real-world, synthetic, and Syn-to-Real samples. The spatial randomization introduced by Syn-Real CutMix further encourages the network to focus on structural cues rather than specific color appearances, thereby improving generalization across domains.

\begin{figure}[t]
    \centering
        \includegraphics[width=3.4in]{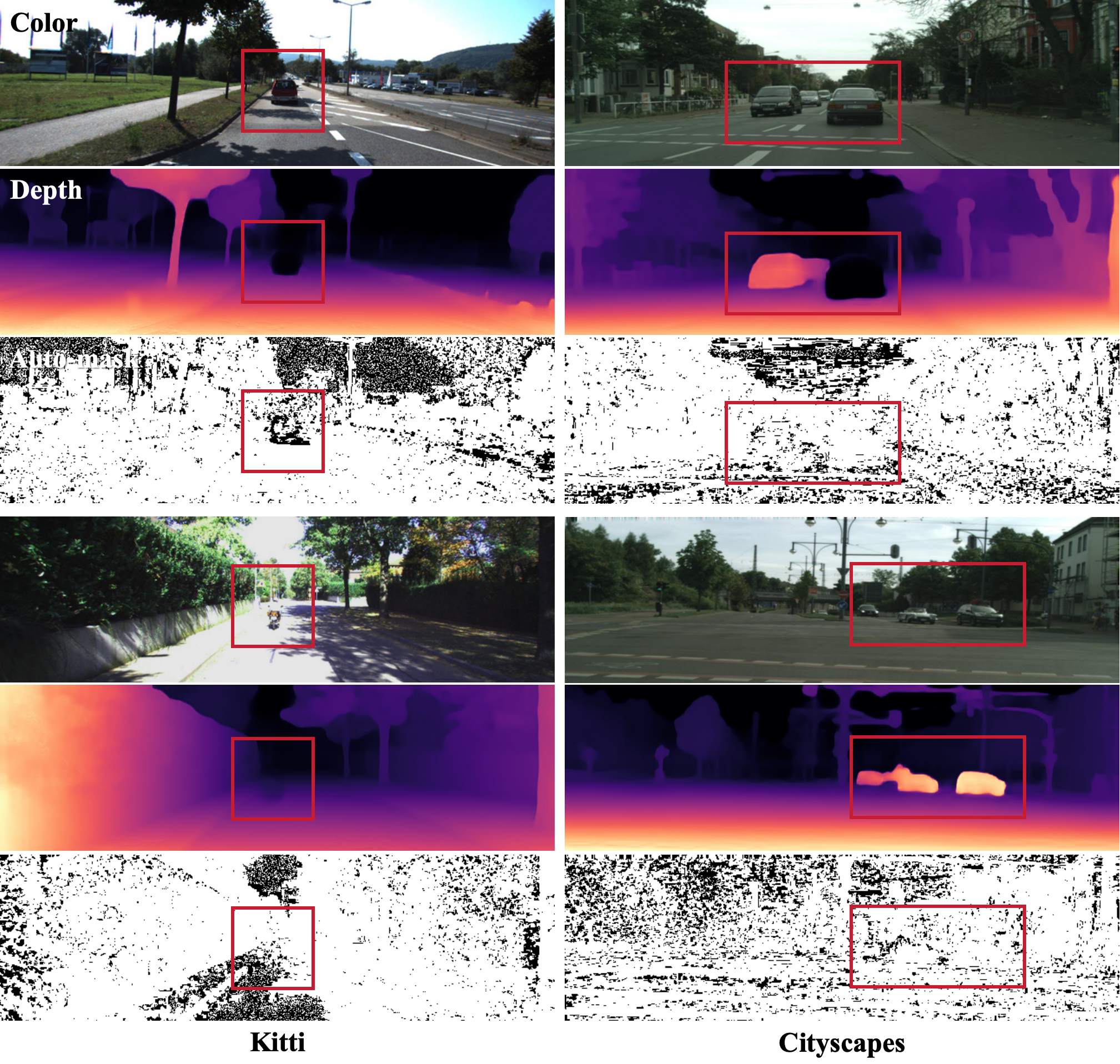}
        \vspace{-3mm}
        \caption{When network capabilities become stronger, and the number of non-rigid objects in the dataset increases, the shortcomings\cite{zhu2024tsudepth} of Auto-masking and Minimum Reprojection Error lead to the reappearance of errors.} 
        \label{fig:fails_AM}
        \vspace{-6mm}
\end{figure}

\begin{figure}[tb]
    \centering
        \includegraphics[width=3.4in]{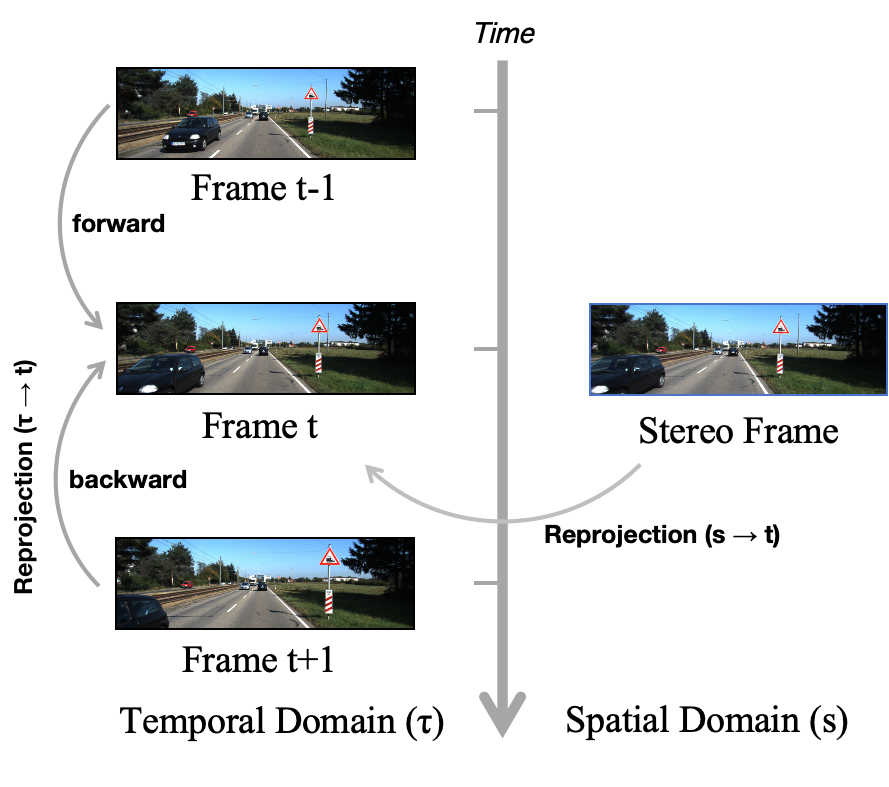}
        \vspace{-3mm}
        \caption{Illustrates the complementary characteristics of temporal and spatial reprojection, which motivates the proposed Auto-UTSF strategy.} 
        \label{fig:Temporal_Spatial}
        \vspace{-5mm}
\end{figure}

\subsection{Auto-learning Uncertainty Temporal--Spatial Fusion}
\label{section:Auto-UTSF}

\par During the evaluation of the Back2Color framework, we observe that the widely adopted auto-masking strategy~\cite{godard2019digging} often fails in complex urban scenes, particularly on the Cityscapes dataset, as illustrated in Fig.~\ref{fig:fails_AM}. By inspecting the auto-masking outputs, we find that non-rigid moving objects, such as vehicles and pedestrians, cannot be reliably filtered. This limitation is further aggravated by the complexity of real-world traffic scenes and the strong fitting capability of modern depth networks, which may compensate for non-rigid motion by adjusting depth predictions to remain consistent with camera poses~\cite{hui2022rm}. Moreover, masking strategies based on optical flow~\cite{zhu2024tsudepth} completely fail for objects moving in the opposite direction of the ego-motion, as shown in Fig.~\ref{fig:fails_AM}.

\par In temporal sequences, freely moving objects violate the rigid-scene assumption underlying depth- and pose-based reprojection, as depicted in Fig.~\ref{fig:Temporal_Spatial}. Auto-masking attempts to identify such pixels by comparing reprojection errors before and after geometric warping:
\begin{equation}
\mu_i = \mathbb{I}\!\left[\min_{k} pe(\mathbf{I}_t,\mathbf{I}_{k\rightarrow t}) 
< \min_{k} pe(\mathbf{I}_t,\mathbf{I}_k)\right],
\label{eq:automask}
\end{equation}
where $k \in \{t-1, t+1\}$. However, this heuristic is insufficient when depth predictions can be adjusted to rationalize non-rigid motion, leading to incorrect supervision.

\par Notably, non-rigid motion only affects temporally adjacent frames, whereas in spatially related stereo pairs captured at the same time, such motion does not occur. Therefore, instead of simply excluding unreliable pixels, introducing spatial (stereo) supervision provides a more direct and effective solution. Nevertheless, temporal and spatial reprojections exhibit complementary characteristics. Temporal reprojection benefits from larger baselines and multi-frame context, making it suitable for distant regions and occlusion completion, while spatial reprojection offers reliable supervision for non-rigid objects but suffers from limited baseline and asymmetric occlusions. These observations motivate a unified strategy that adaptively fuses temporal and spatial supervision.

\par Inspired by uncertainty learning~\cite{zhu2022robust,zhu2024tsudepth,ning2021uncertainty}, we introduce an Auto-learning Uncertainty Temporal--Spatial Fusion (Auto-UTSF) mechanism to combine temporal and spatial reprojection losses. \par In this section, subscripts $t$, $t\!-\!1$, $t\!+\!1$, and $s$ are used to index temporal and stereo frames, while dataset/domain subscripts (e.g., $\mathcal{X}$ and $\mathcal{S}$) are omitted for clarity when the context is unambiguous. And $i$ indexes individual pixels in the image, and $N$ denotes the total number of pixels. Following~\cite{zhu2024tsudepth}, pixel-wise temporal uncertainty logits $\hat{u}^{\tau}_{i,k}$ are predicted for temporally adjacent frames $k \in \{t-1,t+1\}$ and normalized using a softmax operation:
\begin{equation}
u^{\tau}_{i,k} = 1 - \text{Softmax}(\hat{u}^{\tau}_{i,t-1}, \hat{u}^{\tau}_{i,t+1}),
\quad \sum_k u^{\tau}_{i,k} = 1.
\end{equation}
 And The temporal reprojection loss is then defined as
\begin{equation}
\mathcal{L}_{\tau} = \frac{1}{N}\sum_i \sum_k u^{\tau}_{i,k}\,
pe(\mathbf{I}_t,\mathbf{I}_{k\rightarrow t})
+ \alpha \frac{1}{N}\sum_i \sum_k \hat{u}^{\tau}_{i,k},
\label{eq:L_temporal}
\end{equation}
where $\alpha$ is set to $1\times10^{-2}$.

\par To further fuse temporal and spatial supervision, we introduce domain-level uncertainty to balance their contributions. Temporal-domain and spatial-domain uncertainty logits $(\hat{u}^{\tau}_i,\hat{u}^{s}_i)$ are normalized as
\begin{equation}
u^{\tau}_i,\;u^{s}_i = 1 - \text{Softmax}(\hat{u}^{\tau}_i,\hat{u}^{s}_i),
\quad u^{\tau}_i + u^{s}_i = 1.
\end{equation}
The final uncertainty-aware temporal--spatial fusion loss is formulated as
\begin{equation}
\begin{aligned}
\mathcal{L}_{\text{ud}} = \frac{1}{N}\sum_i \Big(
&u^{\tau}_i \sum_k u^{\tau}_{i,k}\,pe(\mathbf{I}_t,\mathbf{I}_{k\rightarrow t}) \\
&+ u^{s}_i\, pe(\mathbf{I}_t,\mathbf{I}_{s\rightarrow t})
\Big) \\
&+ \alpha \frac{1}{N}\sum_i \Big(\sum_k \hat{u}^{\tau}_{i,k}
+ \hat{u}^{\tau}_i + \hat{u}^{s}_i\Big),
\end{aligned}
\label{eq:UTSF}
\end{equation}
where $\mathbf{I}_{s}$ denotes the stereo source image corresponding to the target frame $\mathbf{I}_t$.

\begin{figure}[t]
    \centering
        \includegraphics[width=3.4in]{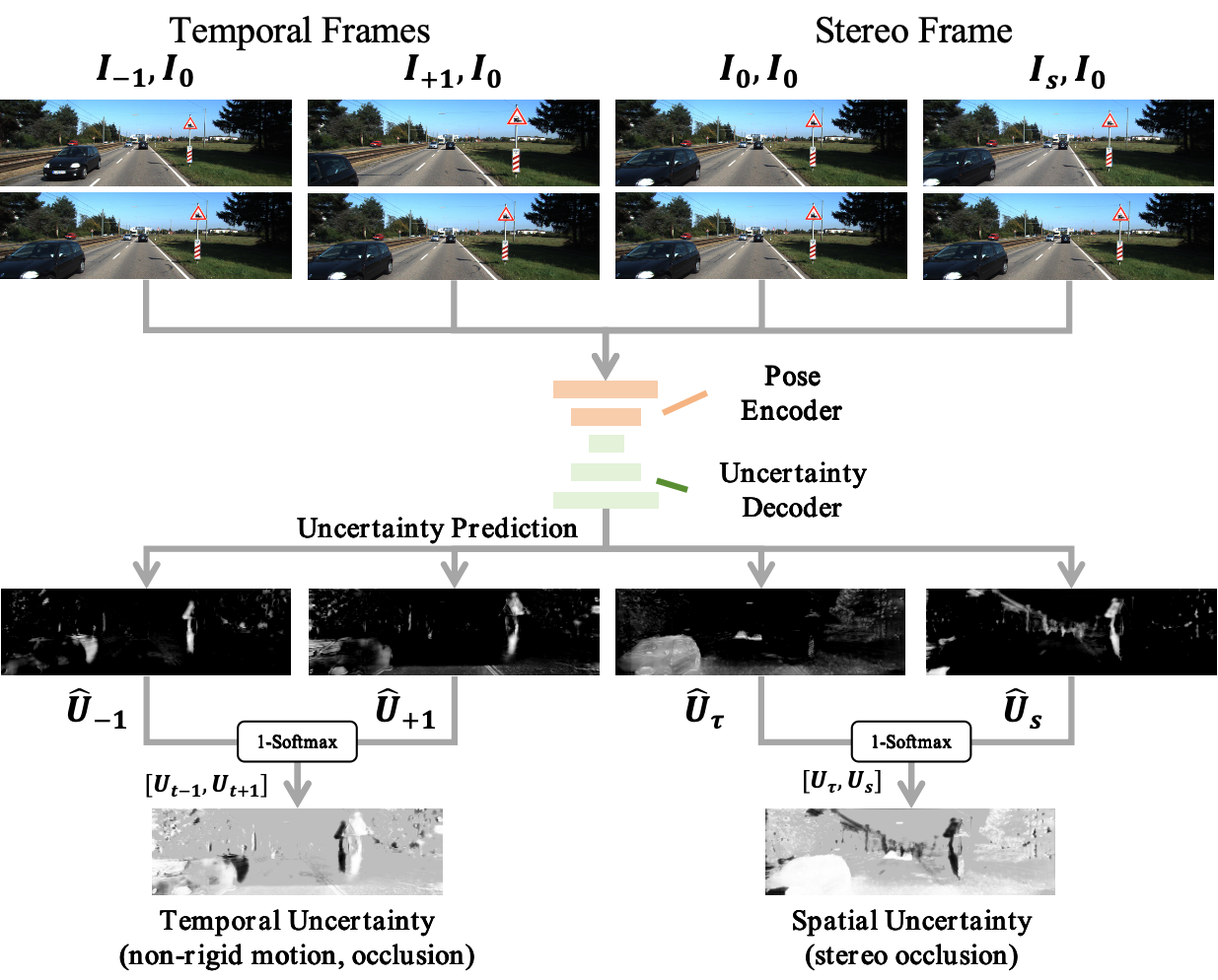}
        \vspace{-3mm}
        \caption{\textbf{Auto-Spatio-Temporal Uncertainty Fusion:} Given temporal and stereo frame pairs, the uncertainty generation module predicts pixel-wise uncertainty logits, which are normalized via a $1-\mathrm{Softmax}$ operation to adaptively weight temporal and spatial reprojection errors.} 
        \label{fig:Uncertainty}
        \vspace{-5mm}
\end{figure}

\par Through uncertainty learning, the model automatically assigns low weights to unreliable regions in each domain. In temporal reprojection, high uncertainty mainly corresponds to occlusions and non-rigid motion, while in spatial reprojection, uncertainty primarily arises from occluded regions caused by viewpoint changes. By fusing uncertainties from both temporal and spatial domains, Auto-UTSF selectively suppresses unreliable supervision and exploits complementary geometric cues, enabling robust depth estimation in dynamic traffic scenes.

\begin{figure}[t]
    \centering
        \includegraphics[width=3.4in]{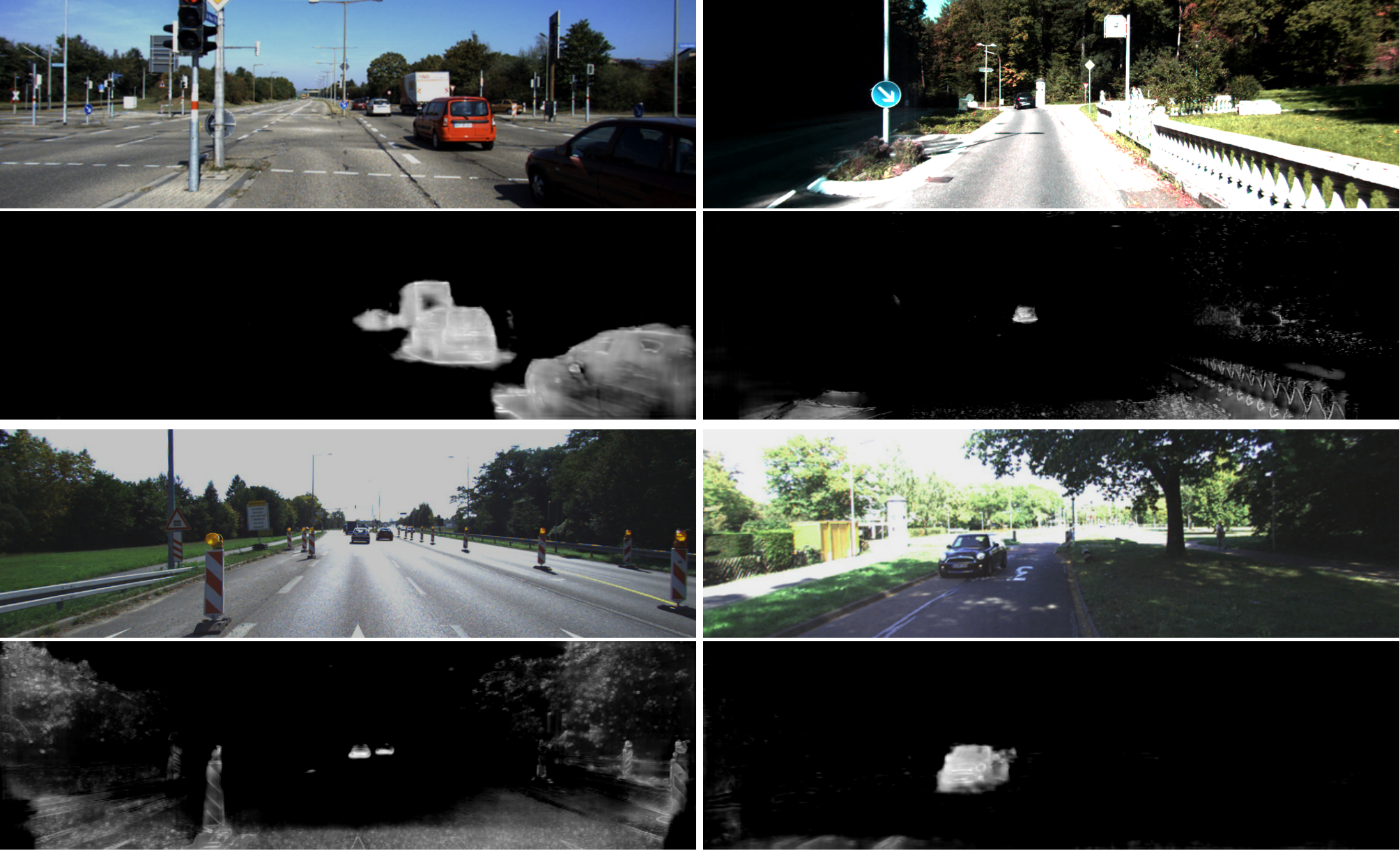}
        \vspace{-3mm}
        \caption{\textbf{Temporal Uncertainty for non-rigid objects:} We can identify non-rigid objects by predicting the uncertainty during the training process, thereby eliminating their impact in the trainning. The fine-grained recognition can even identify very small non-rigid targets.}
        \label{fig:Temporal_Unc}
        \vspace{-5mm}
\end{figure}

\subsection{Training Strategy}

\subsubsection{Synthetic supervised depth loss} 
Since unsupervised depth estimation models based on video sequences cannot recover scale information, we adopt the affine-invariant mean absolute error loss used by MiDas\cite{ranftl2020towards} and Depth Anything\cite{yang2024depth} for joint training with synthesized depth, which has deterministic values varying with different camera parameters:
\begin{equation}
    \mathcal{L}_{sd} = \frac{1}{HW}\sum_i\ |d^*_i -\hat{d}^*_i|,
\end{equation}
where $d^*_i$ and $\hat{d^*_i}$ are the scaled and shifted pixels of predicted $\hat{\textbf{D}}$ and synthetic goundtruth $\textbf{D}$. 
\begin{equation}
    d^*_i = \frac{d_i - median(\textbf{D})}{s(\textbf{D})}, s(\textbf{D}) = \frac{1}{HW}\sum_i |d_i -median(\textbf{D})|.
\end{equation}

\begin{figure}[t]
    \centering
        \includegraphics[width=3.4in]{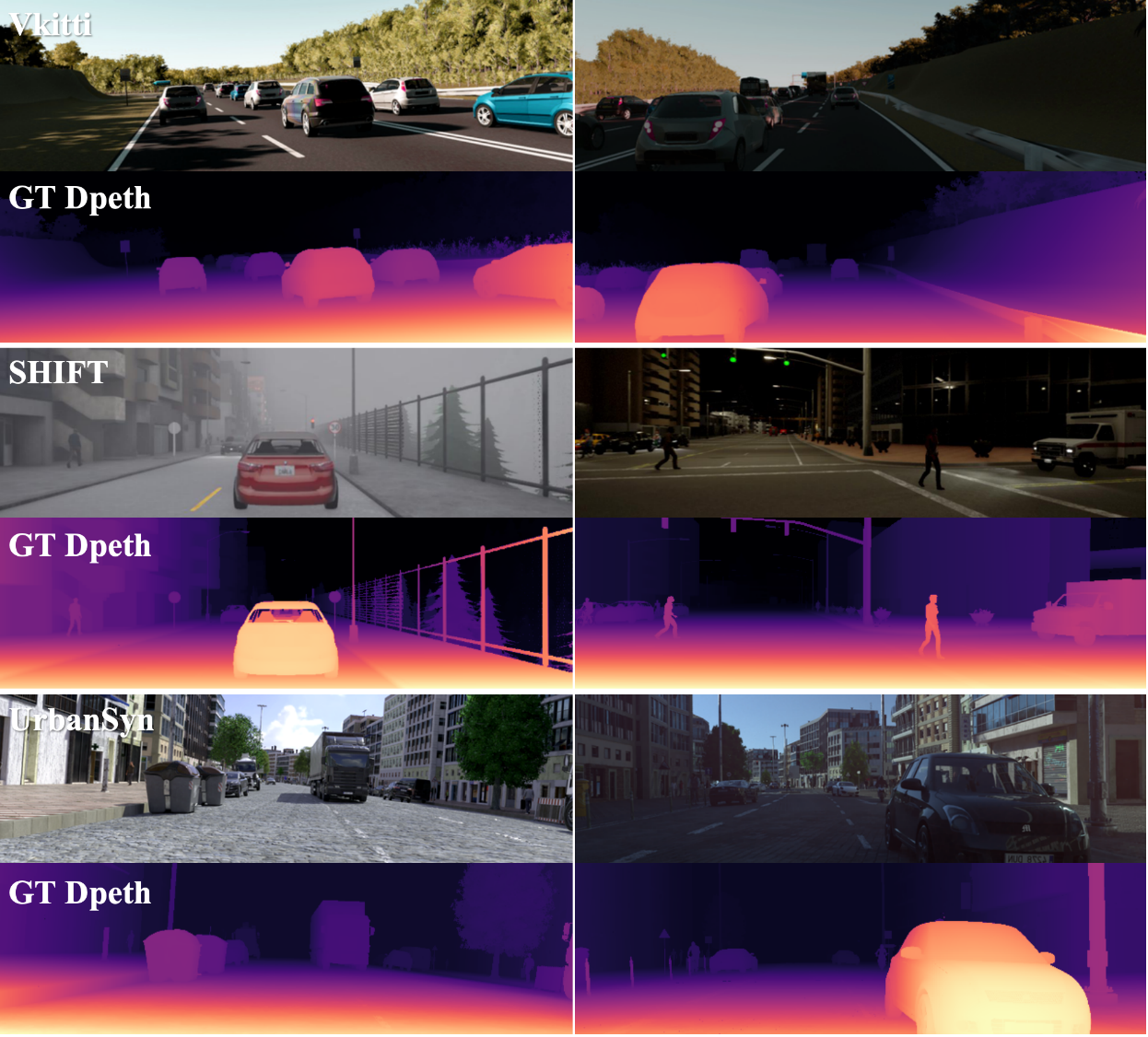}
        \caption{\textbf{The synthetic datasets}: In vKITTI2, depth cannot penetrate transparent entities such as glass windows, whereas in the SHIFT dataset, depth can penetrate transparent entities. SHIFT and UrbanSyn includes simulated pedestrians. And UrbanSyn is the synthetic dataset that most closely resembles real world.} 
        \label{fig:Syn_datasets}
\end{figure}

\begin{figure}[t]
    \centering
        \includegraphics[width=3.4in]{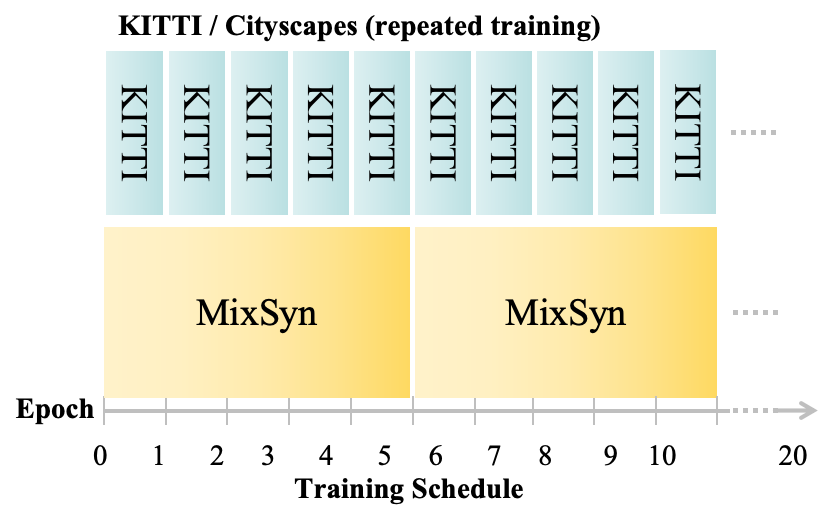}
        \caption{\textbf{The MixSyn dataset} contains more samples than the real-world unsupervised datasets, requiring it to be divided into multiple epochs. While the MixSyn dataset is trained once, the real-world unsupervised datasets may be trained multiple times.}
        \label{fig:Real_Syn}
\end{figure}

\subsubsection{Optimization Strategy}

Swami et al.~\cite{swami2022you} pointed out that, in unsupervised domain adaptation methods, multiple networks and optimization objectives often compete with each other, thereby failing to jointly contribute to effective depth estimation. Such a lack of cooperation among different components makes optimization challenging and limits overall performance. 

To address this issue, we adopt two separate optimizers to decouple the training process and independently optimize the Depth Network on synthetic data and real-world unsupervised data, respectively, as illustrated in Fig.~\ref{fig:framework}. Each training batch contains both real-world samples for unsupervised depth estimation and synthetic color samples.

\par \textbf{Optimizer Step I}: 
The Depth Network is trained in an unsupervised manner on real-world samples using our proposed Auto-UTSF mechanism, while the Color Network is trained in a supervised manner on real-world data to predict color from depth. We set $\gamma = 10^{-3}$, and $\mathcal{L}^{\text{sm}}$ denotes the smoothness loss. The overall loss for this step is defined as
\begin{equation}
    \mathcal{L}_{\mathcal{O}_1} = \mathcal{L}_{\text{d2c}} + \mathcal{L}_{\text{ud}} + \gamma \mathcal{L}^{\text{sm}} .
\end{equation}

\par \textbf{Optimizer Step II}: 
The Depth Network is further trained in a supervised manner using the proposed Syn-Real CutMix samples. The corresponding optimization objective is given by
\begin{equation}
    \mathcal{L}_{\mathcal{O}_2} = \mathcal{L}_{\text{sd}} + \gamma \mathcal{L}^{\text{sm}} .
\end{equation}

\begin{figure*}[t]
    \centering
        \includegraphics[width=7.1in]{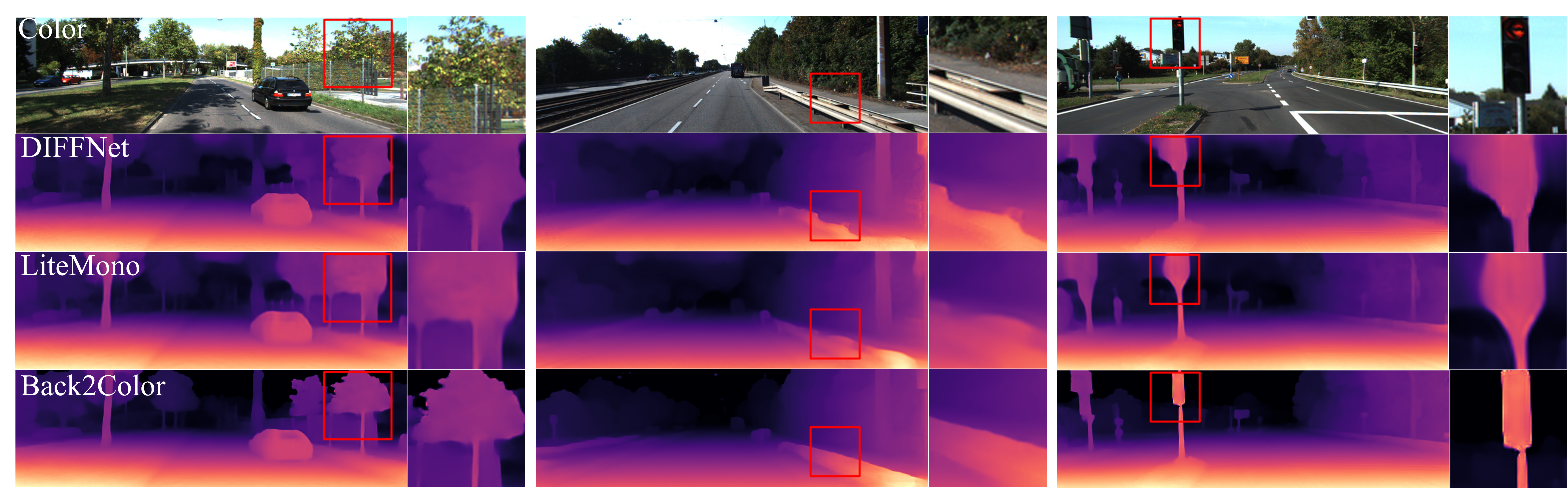}
        \vspace{-5mm}
        \caption{\textbf{Visual of Depth Estimation on KITTI}: Our method yields clearer depth estimates with higher correlation to corresponding colors, sharper boundaries, and fewer areas with chaotic depth clouds.} 
        \label{fig:Depth_show}
        \vspace{-5mm}
    \end{figure*}

\section{Experiments}

In this section, we initially establish the superiority of our Back2Color by comparing it with other state-of-the-art (SoTA) approaches on the KITTI Eigen split and Cityscapes depth estimation benchmark (Tab.\ref{tab:Results_Mono}, Tab.\ref{tab:Results_MS}, and Tab.\ref{tab:Results_CS}). Following this, we conduct ablation experiments to analyze the effects of the proposed methods (Tab.\ref{tab:Abltion_Syn}, Tab.\ref{tab:Results_models} and Tab.\ref{tab:Results_abltion}). Additionally, we compare our approach with other methods that utilize synthetic datasets (Tab.\ref{tab:Results_SynM}), perform comparisons at higher resolutions (Tab.\ref{tab:Results_1280}), and benchmark against the lightest and most effective methods (Tab.\ref{tab:Results_Parameters}). At the same time, we present some visual results of our method on the KITTI and Cityscapes datasets (Fig. \ref{fig:Depth_show}, Fig. \ref{fig:Results_CS}, and Fig. \ref{fig:Results_Syn}).

\subsection{Datasets}
We utilized two unsupervised monocular depth estimation datasets: KITTI\cite{eigen2014depth} and Cityscapes\cite{cordts2016cityscapes}, along with synthetic datasets vKITTI2\cite{cabon2020vKitti2}, SHIFT\cite{shift2022}, and UrbanSyn\cite{gómez2023one}. \textbf{KITTI} contains 39,810 training triplets, 4,424 for validation, and 697 for evaluation, featuring diverse scenes with rich colors and occasional non-rigid objects. \textbf{Cityscapes}, designed for autonomous driving, provides 69,731 monocular triplets from the LeftImg8bit Sequence for training, mainly urban scenes with non-rigid pedestrians and vehicles, often in darker tones. The baseline for KITTI stereo data is 54 cm, while for Cityscapes it is 22 cm. \textbf{vKITTI2}, an upgraded version of Virtual KITTI, uses the Unity game engine for a more photorealistic environment and provides 21,260 frames for training. \textbf{SHIFT} features various weather conditions and times, offers dense depth information with 1 mm resolution, and includes simulated pedestrians. We selected 153,000 frames and synthetic depths from its front-view cameras. \textbf{UrbanSyn}, designed to address the synth-to-real domain gap, provides 7,539 images with depth labels, offering a highly realistic synthetic dataset for various computer vision tasks.
\par To leverage the variety of multiple synthetic datasets, we normalized and shuffled the specifications (size, depth resolution, etc.) of the vKITTI2, SHIFT, and UrbanSyn datasets to create a unified dataset called \textbf{MixSyn}. The MixSyn dataset contains significantly more samples compared to KITTI and Cityscapes. Therefore, during the training process, MixSyn is trained once, while KITTI or Cityscapes is trained for multiple epochs, as shown in Fig.\ref{fig:Real_Syn}. We demonstrate the advantages of using MixSyn in Tab. \ref{tab:Abltion_Syn}.

\begin{table}[]
\centering
\caption{Experiments on the KITTI Eigen Split using only the Monocular Sequence dataset for training.}
\label{tab:Results_Mono}
\resizebox{\columnwidth}{!}{%
\begin{tabular}{@{}cccccccccc@{}}
\toprule
\multirow{2}{*}{\textbf{Method}} & \multirow{2}{*}{\textbf{Sup.\footnotemark[1]}} & \multirow{2}{*}{\textbf{Size}} & \multicolumn{4}{c}{\textbf{Error Metric}}                                 & \multicolumn{3}{c}{\textbf{Accuracy Metric}} \\ \cmidrule(l){4-10} 
                                 &                                    &                                & \textbf{abs\_rel} & \textbf{sq\_rel} & \textbf{rmse} & \textbf{rmse\_log} & \textbf{a1}  & \textbf{a2}  & \textbf{a3}    \\ \midrule
\textbf{Monodepth2\cite{godard2019digging}}              & m                                                            & 192$\times$640                        & 0.115             & 0.903            & 4.863         & 0.193              & 0.877        & 0.959        & 0.981          \\
\textbf{CADepth\cite{yan2021channel}}                 & m                                                           & 192$\times$640                        & 0.105             & 0.769            & 4.535         & 0.181              & 0.892        & 0.964        & 0.983          \\
\textbf{DIFFNet\cite{zhou2021self}}                 & m                                                            & 192$\times$640                        & 0.102             & 0.749            & 4.445         & 0.179              & 0.897        & 0.965        & 0.983          \\
\textbf{MonoFormer\cite{bae2023deep}}              & m                                                           & 192$\times$640                        & 0.106             & 0.839            & 4.627         & 0.183              & 0.889        & 0.962        & 0.983          \\
\textbf{Lite-Mono\cite{zhang2023lite}}               & m                                                            & 192$\times$640                        & 0.101             & 0.729            & 4.454         & 0.178              & 0.897        & 0.965        & 0.983          \\
\textbf{MonoViT\cite{zhao2022monovit}}                 & m                                                            & 192$\times$640                        & 0.099             & 0.708            & 4.372         & 0.175              & 0.900        & 0.967        & \textbf{0.984} \\
\textbf{Back2Color}              & m                                                      & 192$\times$640                        & \textbf{0.096}             & \textbf{0.708}            & \textbf{4.307}         & \textbf{0.173}              & \textbf{0.905}        & \textbf{0.968}        & \textbf{0.984}          \\ \midrule
\textbf{DIFFNet\cite{zhou2021self}}                 & m                                                           & 320$\times$1024                       & 0.097             & 0.722            & 4.345         & 0.174              & 0.907        & 0.967        & 0.984 \\
\textbf{Lite-\cite{zhang2023lite}}               & m                                                            & 320$\times$1024                       & 0.097             & 0.710            & 4.309         & 0.174              & 0.905        & 0.967        & 0.984 \\
\textbf{MonoViT\cite{zhao2022monovit}}                 & m                                                           & 320$\times$1024                       & 0.096             & 0.714            & 4.292         & 0.172              & 0.908        & 0.968        & 0.984 \\
\textbf{Back2Color}              & m                                                      & 320$\times$1024                       & \textbf{0.093}             & \textbf{0.687}            & \textbf{4.171}         & \textbf{0.168}              & \textbf{0.912}        & \textbf{0.970}        & \textbf{0.985}          \\ \bottomrule
\end{tabular}%
}
\end{table}

\footnotetext[1]{\textbf{Sup.}(Supervision) in the tables means what trainning set is used, Video(m),Stereo(s).}

\subsection{Implementation Details}

\par All experiments are implemented using PyTorch~\cite{paszke2019pytorch} and conducted on a single NVIDIA RTX 3090Ti GPU with 24\,GB memory. We adopt the Adam optimizer~\cite{kingma2014adam} with an initial learning rate of $1 \times 10^{-4}$, following the setting of Monodepth2~\cite{godard2019digging}. The learning rate is decayed to $1 \times 10^{-5}$ after 15 epochs, while the remaining hyperparameters are set to $(\beta_1, \beta_2) = (0.9, 0.999)$. The SSIM loss weight is set to $0.85$, and the smoothness regularization weight is set to $\alpha = 1 \times 10^{-3}$. Depth estimation performance is evaluated using standard metrics~\cite{eigen2014depth}, including $\mathrm{AbsRel}$, $\mathrm{SqRel}$, $\mathrm{RMSE}$, $\mathrm{RMSE}_{\log}$, and accuracy thresholds.

\par To ensure computational efficiency and real-time applicability in urban driving scenarios, we adopt a lightweight CNN-based backbone for feature extraction, specifically the Visual Attention Network (VAN)~\cite{guo2023visual}, which captures long-range dependencies with low computational cost. A minor architectural adaptation is introduced to preserve higher-resolution features for dense depth prediction, while keeping the original backbone structure unchanged. The extracted multi-level features are subsequently processed by a high-resolution depth decoder~\cite{he2022ra} to recover fine-grained spatial details and generate depth maps at the input resolution. Experimental results demonstrate that the proposed framework consistently outperforms commonly used CNN-based backbones, such as ResNet-34, HRDepth, and DIFFNet, and achieves competitive performance compared to Transformer-based methods~\cite{zhao2022monovit, zhang2023lite} with significantly lower computational complexity. VAN0-VAN3 denote four variants of the VAN backbone with different parameter scales, as summarized in Table\ref{tab:Results_Parameters}. Although VAN is adopted as the default backbone for efficiency,
Experimental results demonstrate that the proposed framework consistently improves performance when integrated with commonly
used CNN-based backbones, such as ResNet-34, HRDepth, and DIFFNet, and achieves competitive performance compared to Transformer-based
methods~\cite{zhao2022monovit, zhang2023lite} with significantly lower computational complexity, as shown in Fig.\ref{tab:Results_Parameters}.

\begin{table}[]
\centering
\caption{Experiments on the KITTI Eigen Split using only the Monocular Sequence and Stereo datasets for training.}
\label{tab:Results_MS}
\resizebox{\columnwidth}{!}{%
\begin{tabular}{@{}cccccccccc@{}}
\toprule
\multirow{2}{*}{\textbf{Method}} & \multirow{2}{*}{\textbf{Sup.}\footnotemark[1]} & \multirow{2}{*}{\textbf{Size}} & \multicolumn{4}{c}{\textbf{Error Metric}}                                 & \multicolumn{3}{c}{\textbf{Accuracy Metric}} \\ \cmidrule(l){4-10} 
                                  &                                    &                                & \textbf{abs\_rel} & \textbf{sq\_rel} & \textbf{rmse} & \textbf{rmse\_log} & \textbf{a1}   & \textbf{a2}   & \textbf{a3}  \\ \midrule
\textbf{Monodepth2\cite{godard2019digging}}              & ms                                                             & 192$\times$640                        & 0.106             & 0.818            & 4.750         & 0.196              & 0.874         & 0.957         & 0.979        \\
\textbf{HR-Depth\cite{lyu2021hr}}                & ms                                                             & 192$\times$640                        & 0.107             & 0.785            & 4.612         & 0.185              & 0.887         & 0.962         & 0.982        \\
\textbf{CADepth\cite{yan2021channel}}                 & ms                                                             & 192$\times$640                        & 0.102             & 0.752            & 4.504         & 0.181              & 0.894         & 0.964         & 0.983        \\
\textbf{DIFFNet\cite{zhou2021self}}                 & ms                                                             & 192$\times$640                        & 0.101             & 0.749            & 4.445         & 0.179              & 0.898         & 0.965         & 0.983        \\
\textbf{MonoViT\cite{zhao2022monovit}}                 & ms                                                             & 192$\times$640                        & 0.098             & 0.683            & 4.333         & 0.174              & 0.904         & 0.967         & \textbf{0.984}        \\
\textbf{Back2Color}              & ms                                                       & 192$\times$640                        & \textbf{0.092}             & \textbf{0.685}            & \textbf{4.229}         & \textbf{0.170}              & \textbf{0.911}         & \textbf{0.968}         & \textbf{0.984}        \\ \midrule
\textbf{Monodepth2\cite{godard2019digging}}              & ms                                                             & 320$\times$1024                       & 0.106             & 0.806            & 4.630         & 0.193              & 0.876         & 0.958         & 0.980        \\
\textbf{HR-Depth\cite{lyu2021hr}}                & ms                                                             & 320$\times$1024                       & 0.101             & 0.716            & 4.395         & 0.179              & 0.894         & 0.966         & 0.983        \\
\textbf{CADepth\cite{yan2021channel}}                 & ms                                                             & 320$\times$1024                       & 0.096             & 0.694            & 4.264         & 0.173              & 0.908         & 0.968         & 0.984        \\
\textbf{DIFFNet\cite{zhou2021self}}                 & ms                                                           & 320$\times$1024                       & 0.094             & 0.678            & 4.250        & 0.172              & 0.911        & 0.968        & 0.984 \\
\textbf{MonoViT\cite{zhao2022monovit}}                 & ms                                                             & 320$\times$1024                       & 0.093             & 0.671            & 4.202         & 0.169              & 0.912         & 0.969         & 0.984        \\
\textbf{Back2Color}              & ms                                                      & 320$\times$1024                       & \textbf{0.087}             & \textbf{0.644}            & \textbf{4.079}         & \textbf{0.166}              & \textbf{0.920}         & \textbf{0.970}         & \textbf{0.985}        \\ \bottomrule
\end{tabular}%
}
\end{table} 

\begin{table}[]
\centering
\caption{Experiments on the KITTI Eigen Split at 384 $\times$ 1280 resolution.}
\label{tab:Results_1280}
\resizebox{\columnwidth}{!}{%
\begin{tabular}{@{}ccccccccc@{}}
\toprule
\multirow{2}{*}{\textbf{Method}}  & \multirow{2}{*}{\textbf{Sup.}\footnotemark[1]} & \multicolumn{4}{c}{\textbf{Error Metric}}                                 & \multicolumn{3}{c}{\textbf{Accuracy Metric}} \\ \cmidrule(l){3-9} 
                                  &                                       & \textbf{abs\_rel} & \textbf{sq\_rel} & \textbf{rmse} & \textbf{rmse\_log} & \textbf{a1}   & \textbf{a2}   & \textbf{a3}  \\ \cmidrule(r){1-9}
\textbf{PackNet-SfM\cite{guizilini20203d}}                                        & m                                     & 0.107             & 0.802            & 4.538         & 0.186              & 0.889         & 0.962         & 0.981        \\
\textbf{SGDepth\cite{klingner2020self}}                                           & m                                     & 0.107             & 0.768            & 4.468         & 0.186              & 0.891         & 0.963         & 0.982        \\
\textbf{MonoViT\cite{zhao2022monovit}}                                          & m                                     & 0.094             & 0.682            & 4.200         & 0.170              & 0.912         & 0.969         & 0.984        \\
\textbf{PlaneDepth\cite{wang2023planedepth}}                                         & ms                                    & 0.090             & \textbf{0.584}            & 4.130         & 0.182              & 0.896         & 0.962         & 0.981        \\
\textbf{Back2Color}                                       & ms                                    & \textbf{0.086}             & 0.629            & \textbf{4.054}         & \textbf{0.165}              & \textbf{0.920}         & \textbf{0.970}         & \textbf{0.985}        \\ \bottomrule
\end{tabular}%
}
\end{table}

\subsection{Monocular Depth Estimation Performance} 

\begin{figure}[t]
    \centering
        \includegraphics[width=3.4in]{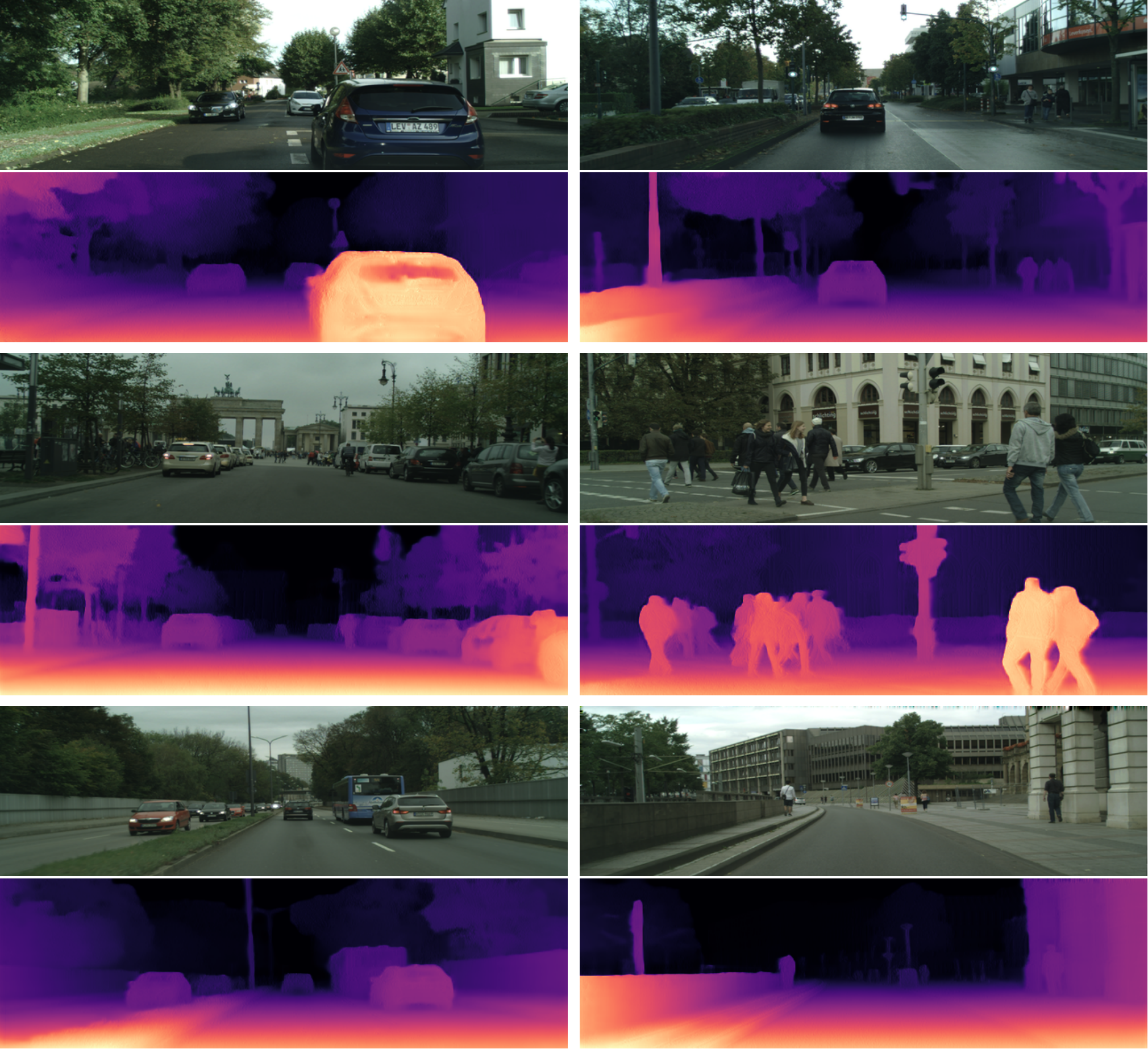}
        \vspace{-3mm}
        \caption{\textbf{The visual results on Cityscapes}: Our method yields clearer depth estimates with higher correlation to corresponding colors, sharper boundaries, and fewer areas with chaotic depth clouds.} 
        \label{fig:Results_CS}
        \vspace{-3mm}
    \end{figure}

Our recommended experimental setup is to use the VAN2 model (balancing efficiency and performance), utilize the MixSyn synthetic dataset, and employ spatiotemporal supervision (ms). If there are no annotations, the results should be based on this setup. The evaluation results are summarized in Tab.\ref{tab:Results_Mono} and Tab.\ref{tab:Results_MS} with KITTI Eigen Split\cite{eigen2014depth} and Tab.\ref{tab:Results_CS} with Cityscapes benchmark, followed by an illustration of their performance qualitatively in Fig.\ref{fig:Depth_show} and Fig.\ref{fig:Results_CS}. Our method outperforms all existing SoTA unsupervised approaches, including some based on transformer\cite{zhang2023lite,zhao2022monovit} with larger params and more complex structures(Tab.\ref{tab:Results_Parameters}). 


\begin{table}[]
\centering
\caption{Experiments on the Cityscapes benchmark at 192$\times$640. Methods* require semantic information, and Back2Color requires stereo pairs during training.}
\label{tab:Results_CS}
\resizebox{\columnwidth}{!}{%
\begin{tabular}{@{}cccccccc@{}}
\toprule
\multirow{2}{*}{\textbf{Method}} & \multicolumn{4}{c}{\textbf{Error Metric}}                                    & \multicolumn{3}{c}{\textbf{Accuracy Metric}}     \\ \cmidrule(l){2-8} 
                                 & \textbf{abs\_rel} & \textbf{sq\_rel} & \textbf{rmse} & \textbf{rmse\_log} & \textbf{a1}    & \textbf{a2}    & \textbf{a3}    \\ \midrule
\textbf{Struct2Depth*\cite{casser2019depth}}            & 0.145             & 1.737             & 7.280          & 0.205               & 0.813          & 0.942          & 0.976          \\
\textbf{GLNet\cite{chen2019self}}                   & 0.129             & 1.044             & 5.361          & 0.212               & 0.843          & 0.938          & 0.976          \\
\textbf{Gordon et al.*\cite{gordon2019depth}}           & 0.127             & 1.330             & 6.960          & 0.195               & 0.830          & 0.947          & 0.981          \\
\textbf{Li et al.\cite{li2021unsupervised}}               & 0.119             & 1.290             & 6.980          & 0.190               & 0.846          & 0.952          & 0.982          \\
\textbf{Lee et al.*\cite{lee2021learning}}              & 0.111             & 1.158             & 6.437          & 0.182               & 0.868          & 0.961          & 0.983          \\
\textbf{RM-Depth\cite{hui2022rm}}                & 0.090             & 0.825             & 5.503          & 0.143               & 0.913          & 0.980          & 0.993          \\ \midrule
\textbf{Back2color}               & \textbf{0.076}    & \textbf{0.784}    & \textbf{5.159} & \textbf{0.127}      & \textbf{0.937} & \textbf{0.984} & \textbf{0.994} \\ \bottomrule
\end{tabular}%
}
\end{table}

\begin{table}[]
\centering
\caption{Experiments with methods using the Synthetic dataset (vKITTI2) at 192×640 resolution and training with monocular sequential only.}
\label{tab:Results_SynM}
\resizebox{\columnwidth}{!}{%
\begin{tabular}{@{}ccccccccc@{}}
\toprule
\multirow{2}{*}{\textbf{Method}} & \multirow{2}{*}{\textbf{Size}} & \multicolumn{4}{c}{\textbf{Error Metric}}                                 & \multicolumn{3}{c}{\textbf{Accuracy Metric}} \\ \cmidrule(l){3-9} 
                                 &                                & \textbf{abs\_rel} & \textbf{sq\_rel} & \textbf{rmse} & \textbf{rmse\_log} & \textbf{a1}   & \textbf{a2}   & \textbf{a3}  \\ \midrule
\textbf{MonoDEVSNet\cite{gurram2021monocular}}             & \textbf{192$\times$640}               & 0.102             & \textbf{0.685}            & \textbf{4.303}         & 0.178              & 0.894         & 0.966         & \textbf{0.984}        \\
\textbf{Back2Color}              & \textbf{192$\times$640}               & \textbf{0.096}             & 0.708            & 4.307         & \textbf{0.173}              & \textbf{0.905}         & \textbf{0.968}         & \textbf{0.984}        \\
\textbf{Swami et al.\cite{swami2022you}}            & \textbf{320$\times$1024}              & 0.103             & \textbf{0.654}            & 4.300         & 0.178              & 0.891         & 0.966         & 0.984        \\ 
\textbf{Back2Color}              & \textbf{320$\times$1024}              & \textbf{0.093}             & 0.687            & \textbf{4.171}         & \textbf{0.168}              & \textbf{0.912}         & \textbf{0.970}         & \textbf{0.985}        \\ \bottomrule
\end{tabular}%
}
\end{table}

\subsection{Ablation Study} 

\begin{table}[]
\centering
\caption{Ablation experiments to test different modules. The experiments use a stereo setup(ms) with a resolution of 320×1024. Row 1 shows the results of Monodepth2\cite{godard2019digging} as a reference.}
\label{tab:Results_abltion}
\resizebox{\columnwidth}{!}{%
\begin{tabular}{lccccccc}
\hline
\multirow{2}{*}{\textbf{Modules}} & \multicolumn{4}{c}{\textbf{Error Metric}}                                     & \multicolumn{3}{c}{\textbf{Accuracy Metric}} \\ \cline{2-8} 
                                  & \textbf{abs\_rel} & \textbf{sq\_rel} & \textbf{rmse} & \textbf{rmse\_log} & \textbf{a1}  & \textbf{a2}     & \textbf{a3} \\ \hline
\textbf{Backbone}                    & 0.094    & 0.703             & 4.435          & 0.173      & 0.909        & 0.969           & 0.984       \\
\textbf{Backbone+MixSyn}                      & 0.095              & 0.724             & 4.672          & 0.172               & 0.910        & 0.969           & 0.984       \\
\textbf{Backbone+MixSyn+B2C}  & 0.090              & 0.688    & 4.320 & 0.169               & 0.915        & \textbf{0.970}  & \textbf{0.985}       \\
\textbf{Backbone+MixSyn+B2C+UTSF}             & \textbf{0.087}              & \textbf{0.644}             & \textbf{4.079}          & \textbf{0.166}               & \textbf{0.920}        & \textbf{0.970}           & \textbf{0.985}       \\ \hline
\end{tabular}%
}
\end{table}

\par As illustrated in Tab. \ref{tab:Results_models}, we conducted experiments on our Back2Color framework by replacing different network structures to demonstrate the effectiveness of our new framework. The results also demonstrated that using a pre-trained checkpoint enhances the network’s ability to process natural images, significantly benefiting depth estimation. Next, we conducted ablation experiments to separately evaluate each of the proposed modules, ad shown in Tab.\ref{tab:Results_abltion} including the MixSyn dataset, the Back2Color training framework, and the auto-learning uncertainty temporal-spatial fusion method (UTSF). The results demonstrated that simply combining real and synthetic data for training did not improve performance. However, applying the Back2Color framework for color domain adaptation effectively addressed this issue. Additionally, Auto-UTSF further improved performance by leveraging advantageous information from both the temporal and spatial domains.
\par Then we compared our method with other noval approaches also joint training with synthetic datasets. Our improvement remains significant in comparison, as shown in Tab.\ref{tab:Results_SynM}.

\begin{table}[!t]
\centering
\caption{Experiments on the KITTI Eigen split at 192×640 resolution. Results show that Back2Color consistently improves performance
across different backbones, indicating that the gain is not solely attributed to backbone selection. VAN* denotes training without ImageNet pretraining.}
\label{tab:Results_models}
\resizebox{\columnwidth}{!}{%
\begin{tabular}{cccccccccc}
\hline
\multirow{2}{*}{\textbf{Method}} & \multirow{2}{*}{\textbf{Model}} & \multicolumn{4}{c}{\textbf{Error Metric}}                                 & \multicolumn{3}{c}{\textbf{Accuracy Metric}} \\ \cline{3-9} 
                                 &                                 & \textbf{abs\_rel} & \textbf{sq\_rel} & \textbf{rmse} & \textbf{rmse\_log} & \textbf{a1}   & \textbf{a2}   & \textbf{a3}  \\ \hline
\textbf{Monodepth2\cite{godard2019digging}}              & Resnet34                        & 0.106             & 0.818            & 4.750         & 0.196              & 0.874         & 0.957         & 0.979        \\
\textbf{Back2Color}              & Resnet34                        & \textbf{0.104}             & \textbf{0.808}            & \textbf{4.617}         & \textbf{0.186}              & \textbf{0.883}         & \textbf{0.962}         & \textbf{0.982}        \\\midrule
\textbf{HRDepth\cite{lyu2021hr}}                 & HRDepth                         & 0.107             & 0.785            & 4.612         & 0.185              & \textbf{0.887}         & 0.962         & 0.982        \\
\textbf{Back2Color}              & HRDepth                         & \textbf{0.105}             & \textbf{0.745}            & \textbf{4.469}         & \textbf{0.182}              & \textbf{0.887}         & \textbf{0.964}         & \textbf{0.984}        \\\midrule
\textbf{DIFFNet\cite{zhou2021self}}                 & DIFFNet                         & 0.101             & 0.749            & 4.445         & 0.179              & 0.898         & 0.965         & 0.983        \\
\textbf{Back2Color}              & DIFFNet                         & \textbf{0.094}             & \textbf{0.660}            & \textbf{4.254}         & \textbf{0.172}              & \textbf{0.904}         & \textbf{0.967}         & \textbf{0.984}        \\\midrule
\textbf{Back2Color}              & VAN2*                           & 0.123             & 0.997            & 4.996         & 0.202              & 0.858         & 0.953         & 0.979        \\
\textbf{Back2Color}              & VAN2                            & \textbf{0.092}             & \textbf{0.685}            & \textbf{4.229}         & \textbf{0.170}              & \textbf{0.911}         & \textbf{0.968}         & \textbf{0.984}        \\ \hline
\end{tabular}%
}
\end{table}

\par Additionally, we used various synthetic datasets such as vKITTI2, SHIFT, and UrbanSyn to real-world unsupervised depth model training, as shown in Tab.\ref{tab:Abltion_Syn}. The results consistently achieved excellent improvement, demonstrating the strong adaptability of our method. To demonstrate that the improvement in our Back2Color method is not due to the increased training samples (Synthetic Datasets) but rather due to the diverse priors brought by various synthetic datasets, we conducted experiments using 20\% and 50\% synthetic samples (Rows 6, 7, and Rows 9, 10) while ensuring the total number of dataset samples remains the same as the KITTI dataset (39,810). The results are consistent with those obtained using the full dataset, proving that the enhancement of our method is not due to the increase in training samples. Additionally, when using the MixSyn mixed synthetic dataset, some metrics (rmse\_log and a1) show significant improvement compared to using a single dataset, demonstrating that our method benefits from learning the variety of multiple synthetic datasets.

Finally, to demonstrate the exceptional efficiency of our model, we compared it with the current best-performing MonoViT\cite{zhao2022monovit} and the lightweight Lite-mono\cite{zhang2023lite}. Our VAN0 model is not only lighter than Lite-mono but also delivers superior performance. Additionally, it achieves competitive results with only one-sixth of the parameters of MonoViT. Meanwhile, our VAN2 model, which has a parameter count similar to MonoViT, shows a significant performance improvement, making it the current state-of-the-art (SOTA) model.

\begin{table*}[]
\centering
\caption{Ablation Study on the impact of Various Synthetic Datasets. }
\label{tab:Abltion_Syn}
\resizebox{\textwidth}{!}{%
\begin{tabular}{@{}ccccccccccccc@{}}
\toprule
\multirow{2}{*}{\textbf{Size}} & \multirow{2}{*}{\textbf{Syn}} & \multicolumn{4}{c}{\textbf{Number of Samples}}                                                                                                         & \multicolumn{4}{c}{\textbf{Error Metric}}                                     & \multicolumn{3}{c}{\textbf{Accuracy Metric}}     \\ \cmidrule(l){3-13} 
&                         & \multicolumn{1}{l}{\textbf{KITTI}} & \multicolumn{1}{l}{\textbf{vKITTI2}} & \multicolumn{1}{l}{\textbf{SHIFT}} & \multicolumn{1}{l}{\textbf{UrbanSyn}} & \textbf{abs\_rel} & \textbf{sq\_rel} & \textbf{rmse} & \textbf{rmse\_log} & \textbf{a1}    & \textbf{a2}    & \textbf{a3}    \\ \midrule
\textbf{192$\times$640}               & \textbf{vKITTI2}              & 39810                           & 21260                             & 0                               & 0                                  & 0.094              & 0.694             & 4.296          & 0.173               & 0.907          & 0.968          & 0.984          \\
\textbf{192$\times$640}               & \textbf{SHIFT}                & 39810                           & 0                                 & 153000                          & 0                                  & \textbf{0.092}     & 0.676             & 4.238          & 0.171               & 0.909          & \textbf{0.969} & 0.984          \\
\textbf{192$\times$640}               & \textbf{UrbanSyn}                & 39810                           & 0                                 & 0                               & 7539                               & 0.093              & \textbf{0.644}    & \textbf{4.198} & 0.172               & 0.907          & 0.968          & 0.984          \\
\textbf{192$\times$640}               & \textbf{MixSyn}               & 39810                           & 21260                             & 153000                          & 7539                               & \textbf{0.092}     & 0.685             & 4.229          & \textbf{0.170}      & \textbf{0.911} & 0.968          & 0.984          \\ \midrule
\textbf{320$\times$1024}              & \textbf{vKITTI2}              & 39810                           & 21260                             & 0                               & 0                                  & 0.090              & 0.677             & 4.173          & 0.169               & 0.914          & 0.969          & 0.984          \\
\textbf{320$\times$1024}              & \textbf{vKITTI2}              & 31848                           & 7962                              & 0                               & 0                                  & 0.090              & 0.675             & 4.143          & 0.168               & 0.917          & 0.970          & 0.985 \\
\textbf{320$\times$1024}              & \textbf{vKITTI2}              & 19905                           & 19905                             & 0                               & 0                                  & 0.091              & 0.684             & 4.192          & 0.169               & 0.913          & 0.969          & 0.985 \\
\textbf{320$\times$1024}              & \textbf{SHIFT}                & 39810                           & 0                                 & 153000                          & 0                                  & 0.089              & 0.651             & 4.158          & 0.167               & 0.916          & 0.970          & 0.985 \\
\textbf{320$\times$1024}              & \textbf{SHIFT}                & 31848                           & 0                                 & 7962                            & 0                                  & 0.090              & 0.684             & 4.158          & 0.168               & 0.916          & 0.970          & 0.984          \\
\textbf{320$\times$1024}              & \textbf{SHIFT}                & 19905                           & 0                                 & 19905                           & 0                                  & 0.090              & 0.680             & 4.144          & 0.167               & 0.915          & 0.970          & 0.985          \\
\textbf{320$\times$1024}              & \textbf{UrbanSyn}                & 39810                           & 0                                 & 0                               & 7539                               & 0.090              & 0.654             & 4.084          & 0.167               & 0.915          & 0.970          & 0.985 \\
\textbf{320$\times$1024}              & \textbf{MixSyn}               & 39810                           & 21260                             & 153000                          & 7539                               & \textbf{0.087}     & \textbf{0.644}    & \textbf{4.079} & \textbf{0.166}      & \textbf{0.920} & 0.970          & 0.985 \\ \bottomrule
\end{tabular}%
}
\end{table*}

\begin{figure}[!t]
    \centering
        \includegraphics[width=3.4in]{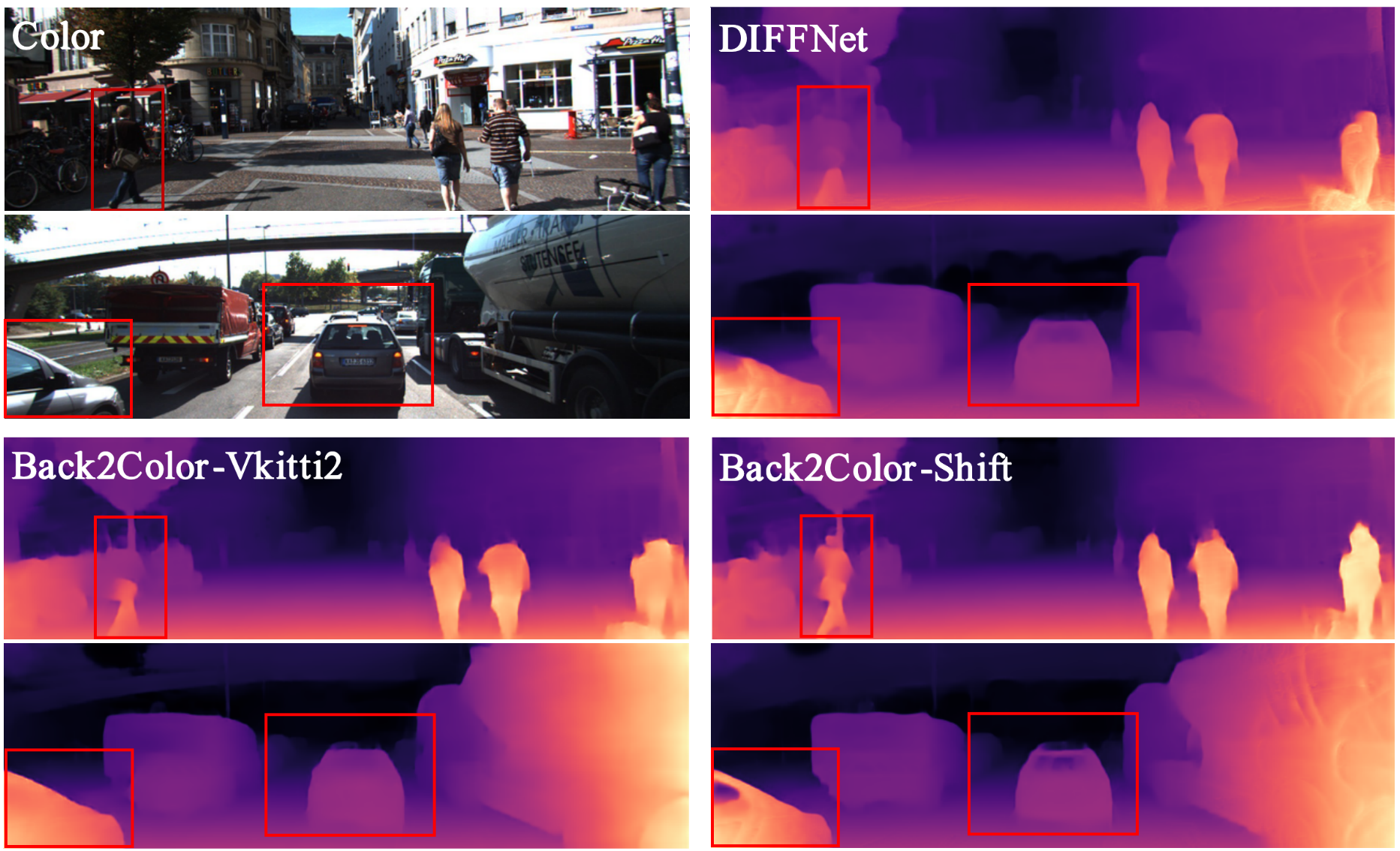}
        \vspace{-3mm}
        \caption{The visual results of joint training with different synthetic datasets. The model trained with SHIFT jointly learns more features of pedestrians while treating glass as transparent entities.} 
        \label{fig:Results_Syn}
        \vspace{-3mm}
    \end{figure}

\begin{table*}[]
\centering
\caption{Ablation study on models with params and computational complexities(GFLOPs).}
\label{tab:Results_Parameters}
\resizebox{\textwidth}{!}{%
\begin{tabular}{cccccccccccc}
\hline
\multirow{2}{*}{\textbf{Method}} & \multirow{2}{*}{\textbf{Model}} & \multirow{2}{*}{\textbf{Size}} & \multirow{2}{*}{\textbf{Params(M)}} & \multirow{2}{*}{\textbf{Flops(G)}} & \multicolumn{4}{c}{\textbf{Error Metric}}                                  & \multicolumn{3}{c}{\textbf{Accuracy Metric}}     \\ \cline{6-12} 
                                 &                                 &                                &                                     &                                    & \textbf{abs\_rel} & \textbf{sq\_rel} & \textbf{rmse}  & \textbf{rmse\_log} & \textbf{a1}    & \textbf{a2}    & \textbf{a3}    \\ \hline
\textbf{MonoViT\cite{zhao2022monovit}}                 & small                           & 320$\times$1024                       & 33.63                               & 159.14                             & 0.093             & 0.671            & 4.202          & 0.169              & 0.912          & 0.969          & \textbf{0.985} \\
\textbf{Lite-mono\cite{zhang2023lite}}               & 8M                              & 320$\times$1024                       & 8.76                                & 29.88                              & 0.097             & 0.710            & 4.309          & 0.174              & 0.905          & 0.967          & 0.984          \\
\textbf{Back2Color}              & VAN0                            & 320$\times$1024                       & 5.17                                & 17.64                              & 0.094             & 0.706            & 4.257          & 0.171              & 0.910          & 0.969          & 0.984          \\
\textbf{Back2Color}              & VAN1                            & 320$\times$1024                       & 18.61                               & 51.48                              & 0.090             & 0.664            & 4.139          & 0.168              & 0.915          & 0.970          & 0.984          \\
\textbf{Back2Color}              & VAN2                            & 320$\times$1024                       & 31.32                               & 76.01                              & \textbf{0.087}    & \textbf{0.644}   & 4.079          & 0.166              & \textbf{0.920} & 0.970          & \textbf{0.985} \\
\textbf{Back2Color}              & VAN3                            & 320$\times$1024                       & 49.5                                & 114.63                             & \textbf{0.087}    & 0.653            & \textbf{4.068} & \textbf{0.165}     & 0.919          & \textbf{0.971} & \textbf{0.985} \\ \hline
\textbf{Back2Color}              & VAN0                            & 384$\times$1280                       & 5.17                                & 17.64                              & 0.093             & 0.676            & 4.169          & 0.171              & 0.911          & 0.969          & 0.984          \\
\textbf{Back2Color}              & VAN1                            & 384$\times$1280                       & 18.61                               & 51.48                              & 0.090             & 0.664            & 4.102          & 0.168              & 0.917          & \textbf{0.970} & 0.984          \\
\textbf{Back2Color}              & VAN2                            & 384$\times$1280                       & 31.32                               & 76.01                              & \textbf{0.086}    & \textbf{0.629}   & \textbf{4.054} & \textbf{0.165}     & \textbf{0.920} & \textbf{0.970} & \textbf{0.985} \\
\textbf{Back2Color}              & VAN3                            & 384$\times$1280                       & 49.5                                & 114.63                             & 0.087             & 0.652            & 4.062          & \textbf{0.165}     & \textbf{0.920} & \textbf{0.970} & 0.984          \\ \hline
\end{tabular}%
}
\end{table*}

\subsection{Visualization Analysis}
\par The impact of synthetic datasets can be analyzed by comparing the joint training with different synthetic datasets, as shown in the Fig.\ref{fig:Results_Syn}. 
\par Firstly, it's undeniable that our method, through joint training with synthetic datasets, achieves clearer boundaries and more detailed depth estimations compared to training solely with unsupervised depth estimation, as depicted in Fig. \ref{fig:Depth_show}. The model successfully captures features of obstacles that unsupervised learning alone struggles to discern. For instance, roadside barriers(Fig.\ref{fig:Depth_show}, Col 2), which run parallel to the direction of vehicle movement and remain unchanged across multiple frames, pose challenges for unsupervised methods but are accurately predicted with the aid of synthetic depths. Synthetic depths are dense and accurate, providing crucial information about these obstacles.
\par Furthermore, the characteristics of synthetic datasets are also learned by the network. For example, as shown in Fig.\ref{fig:Syn_datasets}, the SHIFT dataset has two characteristics compared to the vKITTI2. Firstly, it contains a amount of pedestrians, which are often difficult to estimate correctly in unsupervised depth estimation, typically resulting in partial depth disappearance or blurring, as seen in the estimation results of DIFFNet in the Fig.\ref{fig:Results_CS}, Row 1. When using our method with the SHIFT dataset, the network learns the knowledge of estimating pedestrian, achieving clear and accurate estimation of pedestrian depth, which is impressive.
\par This point can also be inferred from predicted depth of transparent objects, such as glass. As shown in Fig.\ref{fig:Syn_datasets}, in the vKITTI2, transparent objects like glass are treated as solid entities and can be detected, resulting in smooth surfaces in Depth. However, in the SHIFT, transparent objects are considered penetrable targets, leading to discontinuous depth in these regions. This difference in Synthetic dataset definitions is perfectly reflected in the training, as shown in Fig.\ref{fig:Results_Syn}, Row 2. Models trained with the SHIFT dataset recognize and penetrate through vehicle windows, whereas those trained with the vKITTI2 dataset perceive them as flat surfaces.

\begin{figure}[!t]
    \centering
        \includegraphics[width=3.4in]{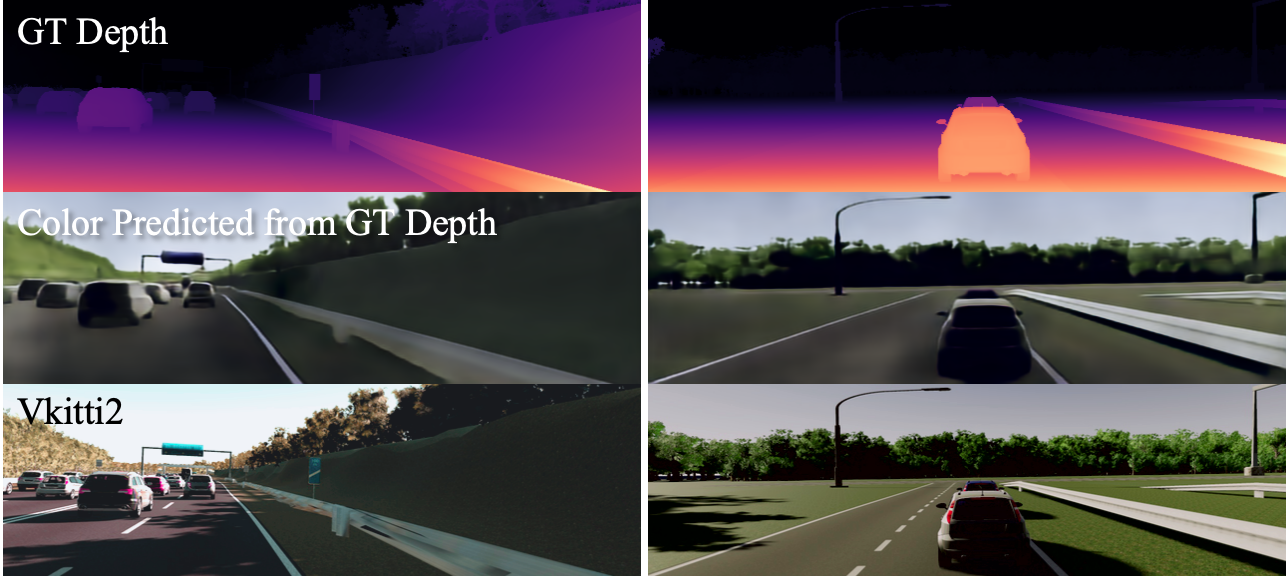}
        \vspace{-3mm}
        \caption{An experiment predicting color from ground truth (GT) depth showed that shadows and road markings vanished, undetectable in the depth image. See Discussion for details.} 
        \label{fig:Discussion_1}
        \vspace{-3mm}
\end{figure}

\subsection{Discussion}
\textbf{Why can the network transform information beyond depth?}
\par In Fig. \ref{fig:examples_back2Color}, Syn-to-Real images exhibit excessive detail surpassing information in depth, such as tree shadows and lane markings. We attribute this to the Color-Depth-Color process, where depth, predicted from color, retains texture details (such as subtle gradient changes at some boundaries). Consequently, the Color Net can recover such gradient changes from the depth map. To validate, we directly trained the network to predict color images from synthesized depth in the synthetic dataset (Fig. \ref{fig:Discussion_1}). The absence of lane markings and shadows in the restored color images confirms our hypothesis.

\par In practice, the ambiguity is significantly reduced by the strong appearance consistency within a target domain (e.g., KITTI or Cityscapes), enabling the depth-to-color network to learn a stable, domain-specific mapping.


\section{Conclusion}
This paper presented Back2Color, an unsupervised monocular depth estimation framework for robust depth perception in urban driving scenarios. By learning a depth-to-color mapping from real-world data and transferring it to synthetic scenes, the proposed method effectively reduces the synthetic-to-real appearance gap and improves depth estimation under real traffic conditions. In addition, an uncertainty-aware temporal-spatial fusion strategy is introduced to handle non-rigid motion and occlusions, leading to more reliable depth predictions in dynamic environments. Experimental results on standard driving benchmarks demonstrate consistent performance improvements with favorable computational efficiency, highlighting the practicality of the proposed approach for vision-based perception in intelligent transportation systems.

\bibliographystyle{IEEEtran}  
\bibliography{main}

@ARTICLE{10321683,
  author={Hu, Yufan and Gao, Junyu and Dong, Jianfeng and Fan, Bin and Liu, Hongmin},
  journal={IEEE Transactions on Multimedia}, 
  title={Exploring Rich Semantics for Open-Set Action Recognition}, 
  year={2024},
  volume={26},
  number={},
  pages={5410-5421},
  doi={10.1109/TMM.2023.3333206}}

@ARTICLE{10130799,
  author={Liu, Hongmin and Jin, Fan and Zeng, Hui and Pu, Huayan and Fan, Bin},
  journal={IEEE Transactions on Neural Networks and Learning Systems}, 
  title={Image Enhancement Guided Object Detection in Visually Degraded Scenes}, 
  year={2023},
  volume={},
  number={},
  pages={1-14},
  doi={10.1109/TNNLS.2023.3274926}}

@article{zhang2024reusable,
  title={Reusable Architecture Growth for Continual Stereo Matching},
  author={Zhang, Chenghao and Meng, Gaofeng and Fan, Bin and Tian, Kun and Zhang, Zhaoxiang and Xiang, Shiming and Pan, Chunhong},
  journal={IEEE Transactions on Pattern Analysis and Machine Intelligence},
  year={2024},
  publisher={IEEE}
}

@article{lin2013absolute,
  title={Absolute depth estimation from a single defocused image},
  author={Lin, Jingyu and Ji, Xiangyang and Xu, Wenli and Dai, Qionghai},
  journal={IEEE Transactions on Image Processing},
  volume={22},
  number={11},
  pages={4545--4550},
  year={2013},
  publisher={IEEE}
}

@inproceedings{li2023temporally,
  title={Temporally consistent online depth estimation in dynamic scenes},
  author={Li, Zhaoshuo and Ye, Wei and Wang, Dilin and Creighton, Francis X and Taylor, Russell H and Venkatesh, Ganesh and Unberath, Mathias},
  booktitle={Proceedings of the IEEE/CVF winter conference on applications of computer vision},
  pages={3018--3027},
  year={2023}
}

@ARTICLE{8016631,
  author={Sheng, Hao and Zhang, Shuo and Cao, Xiaochun and Fang, Yajun and Xiong, Zhang},
  journal={IEEE Transactions on Image Processing}, 
  title={Geometric Occlusion Analysis in Depth Estimation Using Integral Guided Filter for Light-Field Image}, 
  year={2017},
  volume={26},
  number={12},
  pages={5758-5771},
  doi={10.1109/TIP.2017.2745100}}

@ARTICLE{10570231,
  author={Li, Zhenyu and Wang, Xuyang and Liu, Xianming and Jiang, Junjun},
  journal={IEEE Transactions on Image Processing}, 
  title={BinsFormer: Revisiting Adaptive Bins for Monocular Depth Estimation}, 
  year={2024},
  volume={33},
  number={},
  pages={3964-3976},
  doi={10.1109/TIP.2024.3416065}}

@article{kim2018deep,
  title={Deep monocular depth estimation via integration of global and local predictions},
  author={Kim, Youngjung and Jung, Hyungjoo and Min, Dongbo and Sohn, Kwanghoon},
  journal={IEEE transactions on Image Processing},
  volume={27},
  number={8},
  pages={4131--4144},
  year={2018},
  publisher={IEEE}
}

@inproceedings{godard2019digging,
  title={Digging into self-supervised monocular depth estimation},
  author={Godard, Cl{\'e}ment and Mac Aodha, Oisin and Firman, Michael and Brostow, Gabriel J},
  booktitle={Proceedings of the IEEE/CVF international conference on computer vision},
  pages={3828--3838},
  year={2019}
}

@inproceedings{zhou2017unsupervised,
  title={Unsupervised learning of depth and ego-motion from video},
  author={Zhou, Tinghui and Brown, Matthew and Snavely, Noah and Lowe, David G},
  booktitle={Proceedings of the IEEE conference on computer vision and pattern recognition},
  pages={1851--1858},
  year={2017}
}

@inproceedings{godard2017unsupervised,
  title={Unsupervised monocular depth estimation with left-right consistency},
  author={Godard, Cl{\'e}ment and Mac Aodha, Oisin and Brostow, Gabriel J},
  booktitle={Proceedings of the IEEE conference on computer vision and pattern recognition},
  pages={270--279},
  year={2017}
}

@inproceedings{yan2021channel,
  title={Channel-Wise Attention-Based Network for Self-Supervised Monocular Depth Estimation},
  author={Yan, Jiaxing and Zhao, Hong and Bu, Penghui and Jin, YuSheng},
  booktitle={2021 International Conference on 3D Vision (3DV)},
  pages={464--473},
  year={2021},
  organization={IEEE}
}

@inproceedings{lyu2021hr,
  title={Hr-depth: High resolution self-supervised monocular depth estimation},
  author={Lyu, Xiaoyang and Liu, Liang and Wang, Mengmeng and Kong, Xin and Liu, Lina and Liu, Yong and Chen, Xinxin and Yuan, Yi},
  booktitle={Proceedings of the AAAI Conference on Artificial Intelligence},
  volume={35},
  number={3},
  pages={2294--2301},
  year={2021}
}

@inproceedings{hui2022rm,
  title={Rm-depth: Unsupervised learning of recurrent monocular depth in dynamic scenes},
  author={Hui, Tak-Wai},
  booktitle={Proceedings of the IEEE/CVF Conference on Computer Vision and Pattern Recognition},
  pages={1675--1684},
  year={2022}
}

@inproceedings{atapour2018real,
  title={Real-time monocular depth estimation using synthetic data with domain adaptation via image style transfer},
  author={Atapour-Abarghouei, Amir and Breckon, Toby P},
  booktitle={Proceedings of the IEEE conference on computer vision and pattern recognition},
  pages={2800--2810},
  year={2018}
}

@inproceedings{he2022ra,
  title={RA-Depth: Resolution Adaptive Self-Supervised Monocular Depth Estimation},
  author={He, Mu and Hui, Le and Bian, Yikai and Ren, Jian and Xie, Jin and Yang, Jian},
  booktitle={European Conference on Computer Vision},
  pages={565--581},
  year={2022},
  organization={Springer}
}

@article{zhou2021self,
  title={Self-supervised monocular depth estimation with internal feature fusion},
  author={Zhou, Hang and Greenwood, David and Taylor, Sarah},
  journal={arXiv preprint arXiv:2110.09482},
  year={2021}
}

@misc{gómez2023one,
title={All for One, and One for All: UrbanSyn Dataset, the third Musketeer of Synthetic Driving Scenes},
author={Jose L. Gómez and Manuel Silva and Antonio Seoane and Agnès Borrás and Mario Noriega and Germán Ros and Jose A. Iglesias-Guitian and Antonio M. López},
year={2023},
eprint={2312.12176},
archivePrefix={arXiv},
primaryClass={cs.CV}
}

@inproceedings{wang2023planedepth,
  title={PlaneDepth: Self-supervised depth estimation via orthogonal planes},
  author={Wang, Ruoyu and Yu, Zehao and Gao, Shenghua},
  booktitle={Proceedings of the IEEE/CVF Conference on Computer Vision and Pattern Recognition},
  pages={21425--21434},
  year={2023}
}

@inproceedings{guizilini20203d,
  title={3d packing for self-supervised monocular depth estimation},
  author={Guizilini, Vitor and Ambrus, Rares and Pillai, Sudeep and Raventos, Allan and Gaidon, Adrien},
  booktitle={Proceedings of the IEEE/CVF conference on computer vision and pattern recognition},
  pages={2485--2494},
  year={2020}
}

@misc{cabon2020vkitti2,
  title={Virtual KITTI 2},
  author={Cabon, Yohann and Murray, Naila and Humenberger, Martin},
  year={2020},
  eprint={2001.10773},
  archivePrefix={arXiv},
  primaryClass={cs.CV}
}

@InProceedings{shift2022,
    author    = {Sun, Tao and Segu, Mattia and Postels, Janis and Wang, Yuxuan and Van Gool, Luc and Schiele, Bernt and Tombari, Federico and Yu, Fisher},
    title     = {{SHIFT:} A Synthetic Driving Dataset for Continuous Multi-Task Domain Adaptation},
    booktitle = {Proceedings of the IEEE/CVF Conference on Computer Vision and Pattern Recognition (CVPR)},
    month     = {June},
    year      = {2022},
    pages     = {21371-21382}
}

@inproceedings{geiger2012we,
  title={Are we ready for autonomous driving? the kitti vision benchmark suite},
  author={Geiger, Andreas and Lenz, Philip and Urtasun, Raquel},
  booktitle={2012 IEEE conference on computer vision and pattern recognition},
  pages={3354--3361},
  year={2012},
  organization={IEEE}
}

@inproceedings{zhao2020domain,
  title={Domain decluttering: Simplifying images to mitigate synthetic-real domain shift and improve depth estimation},
  author={Zhao, Yunhan and Kong, Shu and Shin, Daeyun and Fowlkes, Charless},
  booktitle={Proceedings of the IEEE/CVF Conference on Computer Vision and Pattern Recognition},
  pages={3330--3340},
  year={2020}
}

@inproceedings{klingner2020self,
  title={Self-supervised monocular depth estimation: Solving the dynamic object problem by semantic guidance},
  author={Klingner, Marvin and Term{\"o}hlen, Jan-Aike and Mikolajczyk, Jonas and Fingscheidt, Tim},
  booktitle={Computer Vision--ECCV 2020: 16th European Conference, Glasgow, UK, August 23--28, 2020, Proceedings, Part XX 16},
  pages={582--600},
  year={2020},
  organization={Springer}
}

@article{gomez2023co,
  title={Co-Training for Unsupervised Domain Adaptation of Semantic Segmentation Models},
  author={G{\'o}mez, Jose L and Villalonga, Gabriel and L{\'o}pez, Antonio M},
  journal={Sensors},
  volume={23},
  number={2},
  pages={621},
  year={2023},
  publisher={MDPI}
}

@inproceedings{zheng2018t2net,
  title={T2net: Synthetic-to-realistic translation for solving single-image depth estimation tasks},
  author={Zheng, Chuanxia and Cham, Tat-Jen and Cai, Jianfei},
  booktitle={Proceedings of the European conference on computer vision (ECCV)},
  pages={767--783},
  year={2018}
}

@inproceedings{huang2018auggan,
  title={Auggan: Cross domain adaptation with gan-based data augmentation},
  author={Huang, Sheng-Wei and Lin, Che-Tsung and Chen, Shu-Ping and Wu, Yen-Yi and Hsu, Po-Hao and Lai, Shang-Hong},
  booktitle={Proceedings of the European Conference on Computer Vision (ECCV)},
  pages={718--731},
  year={2018}
}

@inproceedings{isola2017image,
  title={Image-to-image translation with conditional adversarial networks},
  author={Isola, Phillip and Zhu, Jun-Yan and Zhou, Tinghui and Efros, Alexei A},
  booktitle={Proceedings of the IEEE conference on computer vision and pattern recognition},
  pages={1125--1134},
  year={2017}
}

@article{yang2024depth,
  title={Depth anything: Unleashing the power of large-scale unlabeled data},
  author={Yang, Lihe and Kang, Bingyi and Huang, Zilong and Xu, Xiaogang and Feng, Jiashi and Zhao, Hengshuang},
  journal={arXiv preprint arXiv:2401.10891},
  year={2024}
}

@article{ranftl2020towards,
  title={Towards robust monocular depth estimation: Mixing datasets for zero-shot cross-dataset transfer},
  author={Ranftl, Ren{\'e} and Lasinger, Katrin and Hafner, David and Schindler, Konrad and Koltun, Vladlen},
  journal={IEEE transactions on pattern analysis and machine intelligence},
  volume={44},
  number={3},
  pages={1623--1637},
  year={2020},
  publisher={IEEE}
}

@inproceedings{yun2019cutmix,
  title={Cutmix: Regularization strategy to train strong classifiers with localizable features},
  author={Yun, Sangdoo and Han, Dongyoon and Oh, Seong Joon and Chun, Sanghyuk and Choe, Junsuk and Yoo, Youngjoon},
  booktitle={Proceedings of the IEEE/CVF international conference on computer vision},
  pages={6023--6032},
  year={2019}
}

@inproceedings{swami2022you,
  title={Do What You Can, With What You Have: Scale-aware and High Quality Monocular Depth Estimation Without Real World Labels},
  author={Swami, Kunal and Muduli, Amrit and Gurram, Uttam and Bajpai, Pankaj},
  booktitle={Proceedings of the IEEE/CVF Conference on Computer Vision and Pattern Recognition},
  pages={988--997},
  year={2022}
}

@article{luong1996fundamental,
  title={The fundamental matrix: Theory, algorithms, and stability analysis},
  author={Luong, Quan-Tuan and Faugeras, Olivier D},
  journal={International journal of computer vision},
  volume={17},
  number={1},
  pages={43--75},
  year={1996},
  publisher={Springer}
}

@article{eigen2014depth,
  title={Depth map prediction from a single image using a multi-scale deep network},
  author={Eigen, David and Puhrsch, Christian and Fergus, Rob},
  journal={Advances in neural information processing systems},
  volume={27},
  year={2014}
}

@inproceedings{zhang2023adding,
  title={Adding conditional control to text-to-image diffusion models},
  author={Zhang, Lvmin and Rao, Anyi and Agrawala, Maneesh},
  booktitle={Proceedings of the IEEE/CVF International Conference on Computer Vision},
  pages={3836--3847},
  year={2023}
}

@article{upadhyay2023enhancing,
  title={Enhancing diffusion models with 3d perspective geometry constraints},
  author={Upadhyay, Rishi and Zhang, Howard and Ba, Yunhao and Yang, Ethan and Gella, Blake and Jiang, Sicheng and Wong, Alex and Kadambi, Achuta},
  journal={ACM Transactions on Graphics (TOG)},
  volume={42},
  number={6},
  pages={1--15},
  year={2023},
  publisher={ACM New York, NY, USA}
}

@inproceedings{cordts2016cityscapes,
  title={The cityscapes dataset for semantic urban scene understanding},
  author={Cordts, Marius and Omran, Mohamed and Ramos, Sebastian and Rehfeld, Timo and Enzweiler, Markus and Benenson, Rodrigo and Franke, Uwe and Roth, Stefan and Schiele, Bernt},
  booktitle={Proceedings of the IEEE conference on computer vision and pattern recognition},
  pages={3213--3223},
  year={2016}
}

@article{hochberg1952familiar,
  title={Familiar size and the perception of depth},
  author={Hochberg, Carol Barnes and Hochberg, Julian E},
  journal={The Journal of Psychology},
  volume={34},
  number={1},
  pages={107--114},
  year={1952},
  publisher={Taylor \& Francis}
}

@INPROCEEDINGS{1384826,
  author={Gokturk, S.B. and Yalcin, H. and Bamji, C.},
  booktitle={2004 Conference on Computer Vision and Pattern Recognition Workshop}, 
  title={A Time-Of-Flight Depth Sensor - System Description, Issues and Solutions}, 
  year={2004},
  volume={},
  number={},
  pages={35-35},
  doi={10.1109/CVPR.2004.291}}

@article{li2022progress,
  title={A progress review on solid-state LiDAR and nanophotonics-based LiDAR sensors},
  author={Li, Nanxi and Ho, Chong Pei and Xue, Jin and Lim, Leh Woon and Chen, Guanyu and Fu, Yuan Hsing and Lee, Lennon Yao Ting},
  journal={Laser \& Photonics Reviews},
  volume={16},
  number={11},
  pages={2100511},
  year={2022},
  publisher={Wiley Online Library}
}

@article{blake2011binocular,
  title={Binocular vision},
  author={Blake, Randolph and Wilson, Hugh},
  journal={Vision research},
  volume={51},
  number={7},
  pages={754--770},
  year={2011},
  publisher={Elsevier}
}

@article{ingle2016tesla,
  title={Tesla autopilot: semi autonomous driving, an uptick for future autonomy},
  author={Ingle, Shantanu and Phute, Madhuri},
  journal={International Research Journal of Engineering and Technology},
  volume={3},
  number={9},
  pages={369--372},
  year={2016}
}

@inproceedings{zhu2022robust,
  title={Robust depth completion with uncertainty-driven loss functions},
  author={Zhu, Yufan and Dong, Weisheng and Li, Leida and Wu, Jinjian and Li, Xin and Shi, Guangming},
  booktitle={Proceedings of the AAAI Conference on Artificial Intelligence},
  volume={36},
  number={3},
  pages={3626--3634},
  year={2022}
}

@article{guo2023visual,
  title={Visual attention network},
  author={Guo, Meng-Hao and Lu, Cheng-Ze and Liu, Zheng-Ning and Cheng, Ming-Ming and Hu, Shi-Min},
  journal={Computational Visual Media},
  volume={9},
  number={4},
  pages={733--752},
  year={2023},
  publisher={Springer}
}

@inproceedings{ros2016synthia,
  title={The synthia dataset: A large collection of synthetic images for semantic segmentation of urban scenes},
  author={Ros, German and Sellart, Laura and Materzynska, Joanna and Vazquez, David and Lopez, Antonio M},
  booktitle={Proceedings of the IEEE conference on computer vision and pattern recognition},
  pages={3234--3243},
  year={2016}
}

@inproceedings{huang2018apolloscape,
  title={The apolloscape dataset for autonomous driving},
  author={Huang, Xinyu and Cheng, Xinjing and Geng, Qichuan and Cao, Binbin and Zhou, Dingfu and Wang, Peng and Lin, Yuanqing and Yang, Ruigang},
  booktitle={Proceedings of the IEEE conference on computer vision and pattern recognition workshops},
  pages={954--960},
  year={2018}
}

@inproceedings{dosovitskiy2017carla,
  title={CARLA: An open urban driving simulator},
  author={Dosovitskiy, Alexey and Ros, German and Codevilla, Felipe and Lopez, Antonio and Koltun, Vladlen},
  booktitle={Conference on robot learning},
  pages={1--16},
  year={2017},
  organization={PMLR}
}

@article{haas2014history,
  title={A history of the unity game engine},
  author={Haas, John K},
  journal={Diss. Worcester Polytechnic Institute},
  volume={483},
  number={2014},
  pages={484},
  year={2014}
}

@book{sanders2016introduction,
  title={An introduction to Unreal engine 4},
  author={Sanders, Andrew},
  year={2016},
  publisher={CRC Press}
}

@inproceedings{ganin2015unsupervised,
  title={Unsupervised domain adaptation by backpropagation},
  author={Ganin, Yaroslav and Lempitsky, Victor},
  booktitle={International conference on machine learning},
  pages={1180--1189},
  year={2015},
  organization={PMLR}
}

@article{gurram2021monocular,
  title={Monocular depth estimation through virtual-world supervision and real-world SFM self-supervision},
  author={Gurram, Akhil and Tuna, Ahmet Faruk and Shen, Fengyi and Urfalioglu, Onay and L{\'o}pez, Antonio M},
  journal={IEEE Transactions on Intelligent Transportation Systems},
  volume={23},
  number={8},
  pages={12738--12751},
  year={2021},
  publisher={IEEE}
}

@article{vaswani2017attention,
  title={Attention is all you need},
  author={Vaswani, Ashish and Shazeer, Noam and Parmar, Niki and Uszkoreit, Jakob and Jones, Llion and Gomez, Aidan N and Kaiser, {\L}ukasz and Polosukhin, Illia},
  journal={Advances in neural information processing systems},
  volume={30},
  year={2017}
}

@inproceedings{casser2019depth,
  title={Depth prediction without the sensors: Leveraging structure for unsupervised learning from monocular videos},
  author={Casser, Vincent and Pirk, Soeren and Mahjourian, Reza and Angelova, Anelia},
  booktitle={Proceedings of the AAAI conference on artificial intelligence},
  volume={33},
  number={01},
  pages={8001--8008},
  year={2019}
}

@inproceedings{lee2021learning,
  title={Learning monocular depth in dynamic scenes via instance-aware projection consistency},
  author={Lee, Seokju and Im, Sunghoon and Lin, Stephen and Kweon, In So},
  booktitle={Proceedings of the AAAI Conference on Artificial Intelligence},
  volume={35},
  number={3},
  pages={1863--1872},
  year={2021}
}

@inproceedings{gordon2019depth,
  title={Depth from videos in the wild: Unsupervised monocular depth learning from unknown cameras},
  author={Gordon, Ariel and Li, Hanhan and Jonschkowski, Rico and Angelova, Anelia},
  booktitle={Proceedings of the IEEE/CVF International Conference on Computer Vision},
  pages={8977--8986},
  year={2019}
}

@inproceedings{li2021unsupervised,
  title={Unsupervised monocular depth learning in dynamic scenes},
  author={Li, Hanhan and Gordon, Ariel and Zhao, Hang and Casser, Vincent and Angelova, Anelia},
  booktitle={Conference on Robot Learning},
  pages={1908--1917},
  year={2021},
  organization={PMLR}
}

@inproceedings{liu2021swin,
  title={Swin transformer: Hierarchical vision transformer using shifted windows},
  author={Liu, Ze and Lin, Yutong and Cao, Yue and Hu, Han and Wei, Yixuan and Zhang, Zheng and Lin, Stephen and Guo, Baining},
  booktitle={Proceedings of the IEEE/CVF international conference on computer vision},
  pages={10012--10022},
  year={2021}
}

@inproceedings{chen2019self,
  title={Self-supervised learning with geometric constraints in monocular video: Connecting flow, depth, and camera},
  author={Chen, Yuhua and Schmid, Cordelia and Sminchisescu, Cristian},
  booktitle={Proceedings of the IEEE/CVF International Conference on Computer Vision},
  pages={7063--7072},
  year={2019}
}

@article{dosovitskiy2020image,
  title={An image is worth 16x16 words: Transformers for image recognition at scale},
  author={Dosovitskiy, Alexey and Beyer, Lucas and Kolesnikov, Alexander and Weissenborn, Dirk and Zhai, Xiaohua and Unterthiner, Thomas and Dehghani, Mostafa and Minderer, Matthias and Heigold, Georg and Gelly, Sylvain and others},
  journal={arXiv preprint arXiv:2010.11929},
  year={2020}
}

@inproceedings{ranftl2021vision,
  title={Vision transformers for dense prediction},
  author={Ranftl, Ren{\'e} and Bochkovskiy, Alexey and Koltun, Vladlen},
  booktitle={Proceedings of the IEEE/CVF international conference on computer vision},
  pages={12179--12188},
  year={2021}
}

@inproceedings{bae2023deep,
  title={Deep digging into the generalization of self-supervised monocular depth estimation},
  author={Bae, Jinwoo and Moon, Sungho and Im, Sunghoon},
  booktitle={Proceedings of the AAAI conference on artificial intelligence},
  volume={37},
  number={1},
  pages={187--196},
  year={2023}
}

@article{ning2021uncertainty,
  title={Uncertainty-driven loss for single image super-resolution},
  author={Ning, Qian and Dong, Weisheng and Li, Xin and Wu, Jinjian and Shi, Guangming},
  journal={Advances in Neural Information Processing Systems},
  volume={34},
  pages={16398--16409},
  year={2021}
}

@inproceedings{zhang2023lite,
  title={Lite-mono: A lightweight cnn and transformer architecture for self-supervised monocular depth estimation},
  author={Zhang, Ning and Nex, Francesco and Vosselman, George and Kerle, Norman},
  booktitle={Proceedings of the IEEE/CVF Conference on Computer Vision and Pattern Recognition},
  pages={18537--18546},
  year={2023}
}

@article{varma2022transformers,
  title={Transformers in self-supervised monocular depth estimation with unknown camera intrinsics},
  author={Varma, Arnav and Chawla, Hemang and Zonooz, Bahram and Arani, Elahe},
  journal={arXiv preprint arXiv:2202.03131},
  year={2022}
}

@inproceedings{zhao2022monovit,
  title={Monovit: Self-supervised monocular depth estimation with a vision transformer},
  author={Zhao, Chaoqiang and Zhang, Youmin and Poggi, Matteo and Tosi, Fabio and Guo, Xianda and Zhu, Zheng and Huang, Guan and Tang, Yang and Mattoccia, Stefano},
  booktitle={2022 International Conference on 3D Vision (3DV)},
  pages={668--678},
  year={2022},
  organization={IEEE}
}

@inproceedings{lee2022mpvit,
  title={Mpvit: Multi-path vision transformer for dense prediction},
  author={Lee, Youngwan and Kim, Jonghee and Willette, Jeffrey and Hwang, Sung Ju},
  booktitle={Proceedings of the IEEE/CVF Conference on Computer Vision and Pattern Recognition},
  pages={7287--7296},
  year={2022}
}

@article{kingma2014adam,
  title={Adam: A method for stochastic optimization},
  author={Kingma, Diederik P and Ba, Jimmy},
  journal={arXiv preprint arXiv:1412.6980},
  year={2014}
}

@article{paszke2019pytorch,
  title={Pytorch: An imperative style, high-performance deep learning library},
  author={Paszke, Adam and Gross, Sam and Massa, Francisco and Lerer, Adam and Bradbury, James and Chanan, Gregory and Killeen, Trevor and Lin, Zeming and Gimelshein, Natalia and Antiga, Luca and others},
  journal={Advances in neural information processing systems},
  volume={32},
  year={2019}
}

@article{zhu2024tsudepth,
  title={TSUDepth: Exploring temporal symmetry-based uncertainty for unsupervised monocular depth estimation},
  author={Zhu, Yufan and Ren, Rui and Dong, Weisheng and Li, Xin and Shi, Guangming},
  journal={Neurocomputing},
  pages={128165},
  year={2024},
  publisher={Elsevier}
}

\end{document}